\documentclass{article}

\usepackage{microtype}
\usepackage{graphicx}
\usepackage{booktabs} 
\usepackage{tikz}
\usepackage{algorithm}
\usepackage{algorithmic} 
\usepackage{subcaption}
\usepackage{enumitem}
\usepackage{hyperref}  

\usepackage[accepted]{icml2024}

\usepackage{amsmath}
\usepackage{amssymb}
\usepackage{mathtools}
\usepackage{amsthm}

\usepackage[capitalize,noabbrev]{cleveref}

\theoremstyle{plain}
\newtheorem{theorem}{Theorem}[section]
\newtheorem{proposition}[theorem]{Proposition}

\theoremstyle{definition}
\newtheorem{definition}[theorem]{Definition}

\theoremstyle{remark}

\makeatletter
\renewcommand{\eqref}[1]{\textup{\hyperref[#1]{eq.~\tagform@{\ref*{#1}}}}}
\makeatother

\setlist[itemize]{noitemsep}
\DeclareMathOperator*{\argmin}{arg\,min}

\usepackage{cuted}

\usepackage[textsize=tiny]{todonotes}

\icmltitlerunning{Strongly Isomorphic Neural Optimal Transport Across Incomparable Spaces}

\begin{document}

\twocolumn[
\icmltitle{Strongly Isomorphic Neural Optimal Transport Across Incomparable Spaces}



\icmlsetsymbol{equal}{*}

\begin{icmlauthorlist}
\icmlauthor{Athina Sotiropoulou}{indep}
\icmlauthor{David Alvarez-Melis}{harvard,mrs}
\end{icmlauthorlist}

\icmlaffiliation{indep}{Independent Researcher}
\icmlaffiliation{harvard}{Harvard University}
\icmlaffiliation{mrs}{Microsoft Research, Cambridge, MA}

\icmlcorrespondingauthor{Athina Sotiropoulou}{athina.sotiropoulou@alumni.ethz.ch}

\icmlkeywords{Machine Learning, ICML}

\vskip 0.3in
]



\printAffiliationsAndNotice{}  






\begin{abstract}
Optimal Transport (OT) has recently emerged as a powerful framework for learning minimal-displacement maps between distributions. The predominant approach involves a neural parametrization of the Monge formulation of OT, typically assuming the same space for both distributions. However, the setting across ``incomparable spaces'' (e.g., of different dimensionality), corresponding to the Gromov-Wasserstein distance, remains underexplored, with existing methods often imposing restrictive assumptions on the cost function. In this paper, we present a novel neural formulation of the Gromov-Monge (GM) problem rooted in one of its fundamental properties: invariance to strong isomorphisms. We operationalize this property by decomposing the learnable OT map into two components: (i) an approximate strong isomorphism between the source distribution and an intermediate reference distribution, and (ii) a GM-optimal map between this reference and the target distribution. Our formulation leverages and extends the Monge gap regularizer of \citet{gap_monge} to eliminate the need for complex architectural requirements of other neural OT methods, yielding a simple but practical method that enjoys favorable theoretical guarantees. Our preliminary empirical results show that our framework provides a promising approach to learn OT maps across diverse spaces. 
\end{abstract}

\section{Introduction}
\label{submission}

Transforming samples between distributions lies at the core of machine learning, with applications ranging from generative modeling \citep{gans,rezende2015variational,song2020score} to domain adaptation \citep{courty2017joint} and cell-genomics \citep{bunne2023learning}. Optimal transport (OT) provides an elegant, geometrically-driven approach to address this challenge. Given a source measure $\mu$ supported on a domain $\mathcal{X}$ and a target measure $\nu$ on $\mathcal{Y}$, OT, in its fundamental form, aims at finding a map $T: \mathcal{X} \longrightarrow \mathcal{Y}$, which transports mass from $\mu$ to $\nu$ through the push-forward operation $T \sharp \mu = \nu$, while minimizing a transportation cost \citep{santambrogio2015optimal}.

Most existing work on OT focuses either on defining meaningful distances between distributions \citep{arjovsky2017wasserstein, alvarez2020geometric, pmlr-v97-bunne19a}, or on deriving optimal couplings between discrete sets of unpaired samples \citep{sebbouh2024structured,alvarez2019towards, fickinger2021cross}. In the latter case, the derived coupling cannot generalise to out of distributions samples. To this end, one has to derive the optimal transformation map $T$ in the continuous setting. Yet, its computation on high dimensional settings is notoriously challenging. 

An emerging area of work that aims to address this issue is that of \textit{neural} Optimal Transport, where the OT map $T$ \textit{itself} is parameterized as a neural network, essentially \textit{learning} the \textit{actual} solution $T^*$ to the OT problem \citep{makkuva2020optimal,korotin2023neural_OT,gap_monge,rout2022generative,korotin2020wasserstein2,rout2022generative}. Yet, these methods impose a strict requirement: that the source and target measures are supported on the same space, or at the very least on different but `comparable' spaces (i.e., those across which a meaningful distance can be defined). In particular, they must be of the same dimensionality. This amounts to solving the classic OT formulation of \citet{monge1781memoire}, or the relaxed formulation by \citet{kantorovich2006translocation}, where correspondences are defined as probabilistic couplings. Despite their limitations, these formulations are endowed with the richness of classical optimal transport theory, which is utilized for the design of neural frameworks with robust theoretical guarantees. These (notably) include Brenier's Theorem
\citep{brenier1987decomposition, makkuva2020optimal}, cyclical monotonicity \citep{santambrogio2015optimal, gap_monge} and duality \citep{santambrogio2015optimal, korotin2023neural_OT, rout2022generative, fan2023neural}.

However, transforming samples between distributions that live in \textit{incomparable} spaces, e.g., of different dimensionality or structural form, is essential across a wide range of applications, such as aligning latent representations learned by different models \citep{alvarez2019towards}, or matching samples across different modalities \citep{demetci2022scot, pmlr-v97-bunne19a}. 
The challenge lies in the absence of a meaningful transportation cost across spaces of different dimensionality. 
To this end, the Gromov-Wasserstein (GW) problem \citep{memoli_2011} has been proposed to generalize OT to incomparable spaces. Instead of a cross inter-domain cost $c: \mathcal{X} \times \mathcal{Y} \longrightarrow \mathbb{R}_{+}$, GW computes individual costs $c_{\mathcal{X}}:\mathcal{X}  \times \mathcal{X}\longrightarrow \mathbb{R}_{+}$ and $c_{\mathcal{Y}}:\mathcal{Y}  \times \mathcal{Y}\longrightarrow \mathbb{R}_{+}$ defined within each space and solves a quadratic problem w.r.t the space of couplings. Restricting the space of couplings to an explicit (deterministic) transport map $T$, amounts to the ``hard-assignment'' version of GW, the Gromov-Monge (GM) problem \citep{memoli2022distance}.

Unlike the extensive work for comparable spaces, neural frameworks for the GW and GM problems are largely unexplored. An important challenge in this regard is that the main theorems of classic OT do \textit{not} apply to these versions of the problem. In fact, to the best of our knowledge, the only works in this direction are by \citet{nekrashevich2023neural} and \citet{klein2024entropic}.
\citet{nekrashevich2023neural} restrict their framework for the specific case of using inner product inter-domain costs in GW, solving an equivalent min-max-min problem, while \citet{klein2024entropic} propose a general framework for the entropic GW using neural flow matching.
To the best of our knowledge, Neural frameworks for the GM problem, for general inter-domain cost functions, have yet to be explored.

\textbf{Contributions.} In this work, we propose a \textit{Neural Gromov-Monge} framework, that allows for \textit{any} choice of inter-domain costs. Just as Brenier's Theorem is the theoretical powerhouse of classic neural OT, we utilize a fundamental property of the GM (and GW) problem: its \textit{invariance to strong isomorphisms} \citep{memoli2022distance}. We show that this property can be utilized to estimate the optimal solution map $T^*$ of the GM problem (the GM-optimal map) with neural networks.
Our main contributions are:
\begin{itemize}[itemsep=0pt, parsep=0pt, topsep=0pt, partopsep=0pt]
     \item We show that the solution to the general GM problem between two measures can be decomposed into two maps: an isomorphism and a GM-optimal map (Proposition \ref{prop: gm_mn_rn_equality}, Figure \ref{fig: tripod}).
     \item We show that by parameterizing each map by a neural network, their composition constitutes a universal approximator of \textit{any} transport map between incomparable spaces (Theorem \ref{thm: unviversal approximation}).
     \item We propose a neural algorithm for learning the aforementioned neural composition. By extending the Monge-gap regularizer of \citet{gap_monge} to the GM case, we showcase that if appropriately minimized, our proposed loss is guaranteed to recover true GM-optimal maps. 
     \item We empirically demonstrate that our algorithm can recover known GM-optimal maps on synthetic data. 
\end{itemize}

\section{Preliminaries and Background}

\subsection{Metric measure spaces and Strong Isomorphism}
We consider a compact metric space $(\mathcal{X}, c_{\mathcal{X}})$ endowed with a continuous and measurable metric $c_{\mathcal{X}}: \mathcal{X} \times \mathcal{X} \longrightarrow \mathbb{R}_{+}$. Let $\mu \in \mathcal{P}(\mathcal{X})$ be a Borel probability measure. That is, $\mu$ is fully supported on its domain, i.e $\text{supp}(\mu)=\mathcal{X}$, and  $\mu(\mathcal{X})=1$. Then the triplet $
(\mathcal{X},c_{\mathcal{X}},\mu)$ constitutes a \textit{metric measure space (mm-space)} \citep{memoli_2011}. When it is clear from the context, we will denote $
(\mathcal{X},c_{\mathcal{X}},\mu)$ as simply $\mathcal{X}_{\mu}$. 
Following \citet{vayer_phd}, we denote the space of \textit{all} mm-spaces, with finite $L^p$-size, as $\mathcal{M}_p = \{ \mathcal{X}_{\mu}:= (\mathcal{X},c_{\mathcal{X}},\mu) \; | \; size_{p}(\mathcal{X}_\mu) < +\infty\}$, where $size_{p}(\mathcal{X}_\mu) = \int_{\mathcal{X} \times \mathcal{X}} c_{\mathcal{X}}(x,x')^p  \,d\mu \otimes \,d\mu$ 
, with $\otimes$ denoting the product measure. Let $\mathcal{Y}_{\nu}:=(\mathcal{Y},c_{\mathcal{Y}},\nu)$ be a second mm-space endowed with the Borel probability measure $\nu \in \mathcal{P}(\mathcal{Y})$. To this end, we introduce the concept of \textit{strong isomorphism} between two mm-spaces as:

\begin{definition}\label{def: strong-isomorphism}[Strong Isomorphism] Two mm-spaces $\mathcal{X}_{\mu}\triangleq (\mathcal{X},c_{\mathcal{X}},\mu) \in \mathcal{M}_p$ and $\mathcal{Y}_{\nu}\triangleq(\mathcal{Y},c_{\mathcal{Y}},\nu) \in \mathcal{M}_p $ are strongly isomorphic, denoted as $\mathcal{X}_{\mu}  \cong^{s} \mathcal{Y}_{\nu}$, if there exists a bijective map $\phi: \mathcal{X} \longrightarrow \mathcal{Y}$ s.t:
   \begin{enumerate}[topsep=0pt, parsep=1pt, itemsep=0pt, partopsep=0pt,leftmargin=*]
    \item $\phi$ is an \textit{isometry}, i.e., $\forall x,x'\!\in\!\mathcal{X}$:
    $c_{\mathcal{X}}(x,x') = c_{\mathcal{Y}}(\phi(x),\phi(x'))$.
    \item $\phi$ pushes $\mu$ forward to $\nu$, i.e., $\phi \sharp \mu = \nu$.
   \end{enumerate}
\end{definition}

Note that the push-forward of a measure through a map $\phi$, is the measure $\phi \sharp \mu \in \mathcal{P}(\mathcal{Y})$ satisfying $\phi \sharp \mu(A):=\mu(T^{-1}(A))$ for any measurable set $A \subseteq \mathcal{X}$.

In essence, Definition \ref{def: strong-isomorphism} is a definition of \textit{equivalence}, both from a geometric (condition 1. of \textit{exact} distance preservation) and from a statistical perspective (condition 2. of exact measure preservation).   
Intuitively, the existence of an isometry between the metric spaces $(\mathcal{X},c_{\mathcal{X}})$,$(\mathcal{Y},c_{\mathcal{Y}})$ ensures they have the same spacial configuration. 
On the other hand, the push-forward condition ensures the corresponding mm-spaces have the same probabilistic configuration, i.e the probability mass in both spaces is distributed in exactly the same manner. 
As such, any map $\phi$ that satisfies the conditions in Definition \ref{def: strong-isomorphism}, is a measure preserving isometry, which we will refer to as an \textit{isomorphism}. 
\subsection{The Gromov-Monge problem}\label{sec: gromov-monge}

In this section, we introduce the GM distance between two mm-spaces. Here we consider $\mathcal{X}_{\mu}$ and $\mathcal{Y}_{\nu}$
to be two arbitrary spaces, i.e not necessarily isomorphic. In the general setting, the domains $\mathcal{X}$ and $\mathcal{Y}$ are incomparable. Consider the collection of all possible measure preserving transport maps from $\mathcal{X}_{\mu}$ to $\mathcal{Y}_{\nu}$:
\begin{equation}\label{eq: transport maps mu nu}
    \mathcal{T}(\mu,\nu) =  \{ T: \mathcal{X} \longrightarrow \mathcal{Y} \quad | \quad T \sharp \mu = \nu \}
\end{equation}

The \textit{distortion} (or more precisely, the p-distortion) induced by \textit{any} such map $T \in \mathcal{T}(\mu,\nu)$ is defined as \citep{memoli2022distance}:
\begin{equation}\label{eq: mapping distortion}
    \text{dis}_{p}(T)^p = \iint\limits_{\mathcal{X} \times \mathcal{X}} 
\left| c_{\mathcal{X}}(\mathbf{x}, \mathbf{x'}) - c_{\mathcal{Y}}(T(\mathbf{x}), T(\mathbf{x'})) \right|^p \, d \mu \otimes \mu \\
\end{equation}

Following \citet{memoli2022distance} we can re-write \eqref{eq: mapping distortion} in a more concise manner by using the $L^p$ norm of the function spaces, i.e $\; \text{dis}_{p}(T)^p := \lVert c_{\mathcal{X}} - c_{\mathcal{Y}} \lVert_{L^p(\mu_{T} \otimes \mu_{T})}$. Here, $\; \mu_{T} = (\text{Id} \times T) \sharp \mu\;$ is a measure on $\mathcal{X} \times \mathcal{X}$ and $\text{Id}$ is the identity map on $\mathcal{X}$. We will resort to this form when it is clear from the context. Essentially, \eqref{eq: mapping distortion} captures the degree to which the geometric structure of $\mathcal{X}_{\mu}$ is altered during the (exact) transportation of probability mass from $\mu$ to $\nu$. A lower distortion implies that $T$ preserves the inter-domain distances between points more accurately, i.e is closer to being an isometry, in the p-norm sense. To this end, the Gromov-Monge problem aims at finding the map which induces the least amount of distortion \citep{memoli2022distance}:

\begin{equation}\label{eq: gromov-monge mu nu}
\begin{aligned}
& \text{GM}_{p}(\mu,\nu) = \inf_{T \in \mathcal{T}(\mu,\nu)} \text{dis}_{p}(T) \\
\end{aligned}
\end{equation}
\textbf{GM-optimal maps.} We will refer to the solution of the optimization problem in \eqref{eq: gromov-monge mu nu}, i.e
$\; T^* \in \argmin_{T \in \mathcal{T}(\mu,\nu)} \text{dis}_{p}(T)^{\frac{1}{p}} \; $, as the \textit{GM-optimal map} between the two spaces. 
Intuitively, \eqref{eq: gromov-monge mu nu} tries to match the probability distributions through an exact mapping, whilst also minimizing structural  deformity. Naturally, following Definition \ref{def: strong-isomorphism}, when $\mathcal{X}_{\mu}$ and $\mathcal{Y}_{\nu}$ are strongly isomorphic, we have that $\text{GM}_p(\mu,\nu)=0$ and $T^*$ will be an isomorphism \citep{memoli2022distance}. This means that the GM distance between mm-spaces is \textit{invariant} to strong isomorphisms. In fact, according to \citet{mémoli2022comparison} and \citet{mémoli2022distancedistributionsinverseproblems}, GM defines a Lawvere metric on $\mathcal{M}_p$.

\subsection{The Gromov-Wasserstein Problem}
If instead of optimizing over the set of \textit{exact} correspondences (i.e, \eqref{eq: transport maps mu nu}), we consider \textit{probabilistic} correspondences (i.e, couplings), we get the Gromov-Wasserstein distance as proposed by \citet{memoli_2011}:
\begin{equation}\label{eq: gw mu nu}
\text{GW}_{p}(\mu,\nu) = \inf_{\pi \in \Pi(\mu,\nu)} \text{dis}_{p}(\pi)
\end{equation}
where the infimum is over the set of couplings between $\mu$ and $\nu$, i.e, $\Pi(\mu,\nu) = \{ \pi \in \mathcal{P}(\mathcal{X} \times \mathcal{Y}) \; | \; \pi(A \times Y)=\mu(A) \; ; \; \pi(\mathcal{X} \times B)=\nu(B) \; \text{for any measurable}\; A \subset \mathcal{X}, B \subset \mathcal{Y}\}$ and $\text{dis}_{p}^p(\pi) = \iint\limits_{(\mathcal{X} \times \mathcal{Y})^2} 
\left| c_{\mathcal{X}}(\mathbf{x}, \mathbf{x'}) - c_{\mathcal{Y}}(\mathbf{y}, \mathbf{y'}) \right|^p \, d \pi \otimes \pi$ represents the distortion induced by a coupling $\pi \in \Pi(\mu,\nu)$. 

In essence, the GM problem is the restricted (``hard" assignment) version of the GW problem. The fact that it optimizes over a set of exact point registrations, i.e functions, makes the solution space of GM more suitable for parametrization by neural networks.

%

\textbf{Existence of solutions.} Similar to the Monge problem for comparable spaces \citep{monge1781memoire}, a solution to the GM problem in \eqref{eq: gromov-monge mu nu} might not always exist, i.e, the set of transport maps in \eqref{eq: transport maps mu nu} might be empty. In fact, theoretical guarantees for the existence of GM-optimal maps is an ongoing area of research \citep{vayer_phd, sturm2012, dumont2024existence, salmona2021gromov} and are restricted to very specific conditions that are non-trivial to guarantee in practice. 
In contrast, the Gromov-Wasserstein problem between mm-spaces (i.e, \eqref{eq: gw mu nu}) \textit{always} admits a solution \citep{vayer_phd,memoli_2019}\footnote{Here we are referring to Theorem 12 in \citet{memoli_2019} which considers the GW distance between measure networks (m-nets) but is directly applicable to mm-spaces (which are a specific case of m-nets).}, albeit not necessarily unique 
, i.e, the set of couplings $\Pi(\mu,\nu)$ is always non-empty. Nevertheless, when a solution to \eqref{eq: gromov-monge mu nu}, $T^*$, \textit{does} exist and the measures $\mu$ and $\nu$ are non-atomic, the solutions to the two problems coincide (Theorem 2 in \citet{mémoli2022comparison}), i.e, $\pi^* = (\text{Id},T^*) \sharp \mu$ and $GM(\mu,\nu)=GW(\mu,\nu)$.

\section{GM-Optimal Map Decomposition}

We will henceforth refer to the mm-spaces $\mathcal{X}_{\mu}$ and $\mathcal{Y}_{\nu}$ as the \textit{source} and \textit{target} space respectively and, as in section \ref{sec: gromov-monge}, we consider them to be arbitrary (not necessarily isomorphic). Analogously, we refer to $\mu$ and $\nu$ as the \textit{source} and \textit{target} measures and define their respective supports as compact subsets of Euclidean spaces, i.e $\mathcal{X} \subseteq \mathbb{R}^p$, $\mathcal{Y} \subseteq \mathbb{R}^q$. We assume the general case, where $p \neq q$. Let $\mathcal{Z}_{\rho}:=(\mathcal{Z},c_{\mathcal{Z}},\rho)$ be a third space, with Borel probability measure $\rho \in \mathcal{P}(\mathcal{Z})$ and $\mathcal{Z} \subseteq \mathbb{R}^p$ compact, which we will refer to as the \textit{reference} space. We consider $\mathcal{Z}_{\rho}$ to be \textit{strongly isomorphic} to the source space according to Definition \ref{def: strong-isomorphism}, i.e, $\mathcal{X}_{\mu} \cong^s \mathcal{Z}_{\rho}$. Note that we take the source and reference domains to be of equal dimension $p$.
This constitutes a necessary (albeit not sufficient) condition for strong isomorphism.
 
In this section, we show that the solution of \eqref{eq: gromov-monge mu nu} that directly transports 
$\mu$ onto $\nu$, can be decomposed into a sequence of two maps: an isomorphism from 
$\mu$ to the reference measure
$\rho$,

followed by an optimal transport map from 
$\rho$ to 
$\nu$. 

\subsection{A tripod structure of mm-spaces}

We denote as $\Phi(\mu,\rho)$ the collection of all measure preserving isometries between $\mathcal{X}_{\mu}$ and $\mathcal{Z}_{\rho}$. We now focus on the GM distances (and corresponding GM-optimal maps) between the three mm-spaces. Naturally, between the source and reference we have $\text{GM}(\mu,\rho)=0$, where the corresponding GM-optimal map is an isomorphism $\phi \in \Phi(\mu,\rho)$.


\begin{figure}
    \centering
\begin{tikzpicture}[>=latex]
    \node (X) at (0, 2) {$\mathcal{X}_{\mu}$};
    \node (Y) at (2, 0) {$\mathcal{Y}_{\nu}$};
    \node (Z) at (4, 2) {$\mathcal{Z}_{\rho}$};

    \draw[->] (X) -- (Z) node[midway, above] {$\phi \in \Phi(\mu,\rho)$};
    \draw[->] (Z) -- (Y) node[midway, right=0.5cm] {$\widetilde{T}^* \in \widetilde{\mathcal{T}}(\rho,\nu)$};  
    \draw[->] (X) -- (Y) node[midway, below left] {$T^* \in \mathcal{T}(\mu,\nu)$};
\end{tikzpicture}
    \caption{A tripod structure between mm-spaces where $\mathcal{X}_{\mu} \cong^s \mathcal{Z}_{\rho}$. $\phi \in \Phi(\mu,\rho)$ represents the collection of isomorphisms, while $\mathcal{T}(\mu,\nu)$ and $\widetilde{\mathcal{T}}(\rho,\nu)$ are the sets of all transport maps between the corresponding spaces.}
    \label{fig: tripod}
\end{figure}
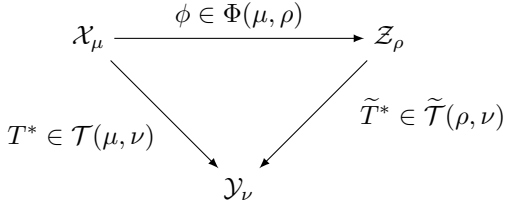

Now let us focus on the following tripod structure: \[ \mathcal{X}_{\mu} \overset{\mathcal{T}(\mu,\nu)}{\xrightarrow{\hspace*{1cm}}} \mathcal{Y}_{\nu} \overset{\widetilde{\mathcal{T}}(\rho,\nu)}{\xleftarrow{\hspace*{1cm}}} \mathcal{Z}_{\rho}\]
where $\widetilde{\mathcal{T}}(\rho,\nu)$ is the collection of all transport maps between $\mathcal{Z}_{\rho}$ and $\mathcal{Y}_{\nu}$:
\begin{equation}\label{eq: transport maps rho nu}
    \widetilde{\mathcal{T}}(\rho,\nu) = \{\widetilde{\mathcal{T}}: \mathcal{Z} \longrightarrow \mathcal{Y} \; | \; \widetilde{\mathcal{T}} \sharp \rho = \nu \}
\end{equation}

and $\mathcal{T}(\mu,\nu)$ is given by \eqref{eq: transport maps mu nu}. Subsequently, the GM problem $\text{GM}_p(\mu,\nu)$ between the source and target spaces is given by \eqref{eq: gromov-monge mu nu}. The GM distance between $\mathcal{Z}_{\rho}$ and $\mathcal{Y}_{\nu}$ will then be defined as:
\begin{equation}\label{eq: gm rho nu}
\text{GM}_{p}(\rho,\nu) = \inf_{\widetilde{T} \in \widetilde{\mathcal{T}}(\rho,\nu)} \text{dis}_{p}(\widetilde{T}) 
\end{equation}

where $\; \text{dis}_{p}(\widetilde{T})^p := \lVert c_{\mathcal{Z}} - c_{\mathcal{Y}} \lVert_{L^p(\rho_{\widetilde{T}} \otimes \rho_{{\widetilde{T}}})}$, with $\; \rho_{\widetilde{T}} = (\text{Id} \times \widetilde{T}) \sharp \rho  \in \mathcal{P}(\mathcal{Z} \times \mathcal{Z})$. We denote as $\widetilde{T}^* \in \widetilde{\mathcal{T}}(\rho,\nu)$, the solution to 
\eqref{eq: gm rho nu}.
It can be easily shown that the GM problems \eqref{eq: gromov-monge mu nu} and \eqref{eq: gm rho nu} are equivalent. Formally, we introduce the following Proposition, which is a consequence of the isomorphic invariance of the GM problem, as introduced in section \ref{sec: gromov-monge}. 

\begin{proposition}\label{prop: gm_mn_rn_equality}
Let $\mathcal{X}_{\mu},\mathcal{Z}_{\rho}$ ,$\mathcal{Y}_{\nu} \in \mathcal{M}_p$ such that $\mathcal{X}_{\mu}  \cong^{s} \mathcal{Z}_{\rho}$. Then for $p \in \left[ 1, \infty \right)$ it holds that $GM_p(\mu,\nu)=GM_p(\rho,\nu)$.
\end{proposition}

We provide the proof of Proposition \ref{prop: gm_mn_rn_equality} in Appendix \ref{app: proof prop gm equal}.

\subsection{Optimal map decomposition}

Now let us focus on the transport problem from the source to the target space, i.e, $\mathcal{X}_{\mu} \longrightarrow \mathcal{Y}_{\nu}$. Consider the  structure illustrated in Figure \ref{fig: tripod}. Instead of transporting mass directly from $\mathcal{X}_{\mu}$ to $\mathcal{Y}_{\nu}$,  
we introduce a ``detour", by first mapping $\mu$ onto $\rho$ through an isomorphism $\phi \in \Phi(\mu,\rho)$ and subsequently mapping $\rho$ onto $\nu$ through a transport map $\widetilde{T} \in \widetilde{\mathcal{T}}(\rho,\nu)$, i.e, $\mathcal{X}_{\mu} \overset{{\phi}} \longrightarrow \mathcal{Z}_{\rho} \overset{\widetilde{T}}\longrightarrow \mathcal{Y}_{\nu}$. This alternate mapping can be expressed, naturally, as the composition $\widetilde{T} \circ \phi : \mathcal{X} \longrightarrow \mathcal{Y}$. It is straightforward to show that since $\mathcal{X}_{\mu} \cong^s \mathcal{Z}_{\rho}$, i.e $\phi \sharp \mu=\rho$, we have
$(\widetilde{T} \circ \phi) \sharp \mu = \widetilde{T}\sharp (\phi \sharp \mu)=\widetilde{T}\sharp \rho = \nu$.
As such, we can define the collection of all such composition maps as: 
\begin{align}\label{eq: transport map constrained}
\begin{split}
    \mathcal{I(\mu, \nu)} = \{ T: \mathcal{X} \to \mathcal{Y} \mid T \sharp \mu = \nu \;,\; \\
    T \triangleq \widetilde{T} \circ \phi \;,\; \phi \in {\Phi}(\mu,\rho) \; , \; \widetilde{T} \in \mathcal{\widetilde{T}}(\rho, \nu) \} 
\end{split}
\end{align}
where $\mathcal{I}(\mu,\nu) \subseteq \mathcal{T}(\mu,\nu)$.
Since we restrict the collection of transport maps between the source and target space to only those that can be decomposed as in \eqref{eq: transport map constrained},
the proposed GM problem becomes a \textit{constrained} version of \eqref{eq: gromov-monge mu nu}: 

\begin{align}\label{eq:gm-constraint-mn}
\text{CGM}_p(\mu,\nu) = 
    \inf_{T \in \mathcal{I}(\mu,\nu)}  \text{dis}_p(T)
\end{align}

where $\text{dis}_p(T)$ is given by \eqref{eq: mapping distortion}. Note that since the isomorphism $\phi \in \Phi(\mu,\rho)$ is already a solution (albeit not necessarily unique) to the $\text{GM}(\mu,\rho)=0$ problem,  it is considered \textit{fixed}. As such, we do not need to use a double infinum in \eqref{eq:gm-constraint-mn}. In other words, given \textit{any} isomorphism between the source and target space, we only need to optimize over the set $\mathcal{\widetilde{T}}(\rho,\nu)$, i.e finding the GM-optimal map $\widetilde{T}$ what pushes $\mu$ onto $\phi \sharp \mu $. To this end, \eqref{eq:gm-constraint-mn} can be re-written as:

\begin{equation}\label{eq:gm-constraint-mn-reframed} 
\text{CGM}_p(\mu,\nu) = 
    \inf_{\widetilde{T} \sharp \rho = \nu}  \text{dis}_p(\widetilde{T} \circ \phi) 
\end{equation}

where $\; \text{dis}_p(\widetilde{T} \circ \phi) = \lVert c_{\mathcal{X}} - c_{\mathcal{Y}} \lVert_{L^p(\mu_{\widetilde{T} \circ \phi} \otimes \mu_{\widetilde{T} \circ \phi})}$ with $ \; \mu_{\widetilde{T} \circ \phi} = (\text{Id},\widetilde{T} \circ \phi)$. Then, using Proposition \ref{prop: gm_mn_rn_equality} we show the following result.

\begin{proposition}\label{prop: cgm=gm}
    Assume problem \eqref{eq: gromov-monge mu nu} admits at least one solution, i.e $\exists \; T^* \in \argmin_{T \in  \mathcal{T}(\mu,\nu)} \text{dis}_p(T)$.
    Then, given a reference space s.t $\mathcal{X}_{\mu} \cong^s \mathcal{Z}_{\rho}$ and any isomorphism $\phi \in \Phi(\mu,\rho)$, there exists ${\widetilde{T}}^* \in \argmin_{\mathcal{\widetilde{T}}(\rho,\nu)} \text{dis}_p({\widetilde{T}})$ s.t 
    the composition map ${\widetilde{T}}^* \circ 
    \phi$ is also a solution to \eqref{eq: gromov-monge mu nu}. That is, 
    the optimization problems \eqref{eq: gromov-monge mu nu} and \eqref{eq:gm-constraint-mn-reframed} are equivalent, i.e, $\text{GM}_p(\mu,\nu)=\text{CGM}_p(\mu,\nu)$.
\end{proposition}

We provide the proof of Proposition \ref{prop: cgm=gm} in Appendix \ref{app: cgm=gm}.

Proposition \ref{prop: cgm=gm}, shows that we 
can decompose the search for an optimal solution $T^* \in \mathcal{T}(\mu,\nu)$ to $\text{GM}_p(\mu,\nu)$ into the search for an isomorphism $\phi \in \Phi(\mu,\rho)$ and a map ${\widetilde{T}}^* \in \mathcal{\widetilde{T}}(\rho,\nu)$. It shows that we can effectively break the complex problem of optimally transporting one measure onto another into two structured sub-problems, without losing GM-optimality. Note that, assuming non-unique optimality, the above result does not necessarily mean that \textit{every} GM-optimal map of \eqref{eq: gromov-monge mu nu} can be decomposed in this manner but that at least \textit{some} are, which is sufficient for our analysis. 

\textbf{Geometric Intuition.} From a geometric perspective, we can interpret Proposition \ref{prop: cgm=gm} as follows: Any exact transportation of probability mass that aims to minimize distortion from one mm-space onto another, can be represented as a sequence of two geometric transformations. An isomorphic transformation, which in Euclidean space includes rotations reflections and translations and a geometric ``deformation" which represents the distortion of the initial geometry of the source metric space, e.g shearing. This two-stage process highlights the interplay in GM-optimal transport between maintaining intrinsic geometric properties and adapting to new probabilistic configurations. 

\section{Isomorphism Invariant Neural Gromov-Monge}

In this section, we introduce our neural framework for approximating the GM-optimal composition map $\widetilde{T}^* \circ \phi$ in Proposition \ref{prop: cgm=gm} for $p=2$. We use neural networks $\phi_{\omega}: \mathcal{X} \longrightarrow \mathcal{Z}$ and $\widetilde{T}_{\theta}: \mathcal{Z} \longrightarrow \mathcal{Y}$ to parameterize  $\phi$ and $\widetilde{T}^*$ respectively. For convenience, we will refer to $\phi_{\omega}$ as the \textit{isomorphism network} and to $\widetilde{T}_{\theta}$ as the \textit{transport network}.
To this end, in section \ref{sec: nerual_approximation}, we prove the following theoretical result: that the composition of neural networks $\widetilde{T}_{\theta} \circ \phi_{\omega}$ can approximate \textit{any} transport map  between a source and a target measure. Subsequently, in section \ref{sec: gromov-monge}, we present our proposed learning procedure for approximating GM-optimal maps, s.t $\widetilde{T}_{\theta} \circ \phi_{\omega} \approx \widetilde{T}^* \circ \phi$.

\subsection{Neural Network Compositions as Universal Approximators of Transport Maps}\label{sec: nerual_approximation}

Our analysis is inspired by Theorem 1 in \citet{korotin2023neural_OT}, which states that neural networks can approximate any stochastic transport map in the $L^2$ norm. 

\textbf{Transport maps in the $L^p$ space.} 
Note that the set of transport \textit{maps} between mm-spaces can include both continuous \textit{and} discontinuous functions. More formally, let $L^2_{\mu}(\mathcal{X},\mathcal{Y})$ be the space of quadratically integrable functions w.r.t $\mu$, i.e $\{ f: \mathcal{X} \longrightarrow \mathcal{Y} \; | \; \lVert f \lVert_{L^2(\mu)} < +\infty\}$, which is known to include both continuous and irregular functions. Assuming $\mu$ has a finite second moment, it can be shown that, for any transport map $T$ from $\mathcal{X}_{\mu}$ to $\mathcal{Y}_{\nu}$, we have $T \in L^2_{\mu}(\mathcal{X},\mathcal{Y})$ \citep{korotin2023neural_OT}. As such, 
given a neural network $\phi_{\omega}$ which is \textit{specifically modeled to approximate an isometry}, we aim to show that $\widetilde{T}_{\theta} \circ \phi_{\omega}$ is dense in the function space $L^2_{\mu}(\mathcal{X},\mathcal{Y})$. We base our analysis on the result by \citet{kratsios2020non} who show that the composition of a continuous 
injective map $\phi$ with a ReLU neural network is dense in the space of continuous functions. In our case, since $\phi$ is an isomorphism, it inherently satisfies both continuity and injectivity (note that an isomorphism is also bijective). We present our result in the following Theorem. 
\begin{theorem}\label{thm: unviversal approximation}
    Let $T \in \mathcal{T}(\mu,\nu)$ be a transport map between $\mu$ and $\nu$, where $\nu$ has finite second moment. 
    Then, there exists a feed-forward  
    ReLU neural network 
    $\widetilde{T}_{\theta}: \mathcal{Z} \longrightarrow  \mathcal{Y}$
    and a neural network 
    $\phi_{\omega}: \mathcal{X} \longrightarrow  \mathcal{Z}$ 
    with any nonaffine continuous activation function which is continuously
differentiable at at least one point, such that, $\forall \epsilon > 0$ we have:

\begin{equation}
    \lVert T- \widetilde{T}_{\theta} \circ \phi_{\omega} \lVert_{L^2_{\mu}} \leq \epsilon
\end{equation}
\end{theorem}

We provide the full proof of Theorem \ref{thm: unviversal approximation} in Appendix \ref{app: proof theorem universal}.

\textbf{Intuition behind Theorem \ref{thm: unviversal approximation}}. In \citet{kratsios2020non}, they consider $\phi$ to be a (continuous and injective) \textit{feature map} from a (possibly) non-Euclidean to a Euclidean feature space. Their result shows that the presence of such a map, in the initial layer of an architecture, does not compromise its approximation capabilities. Our result can be viewed in a similar fashion. An isomorphism $\phi \in \Phi(\mu,\nu)$ can be viewed as a `feature map' from the source $\mathcal{X}_{\mu}$ to the reference space $\mathcal{Z}_{\rho}$. To this end, it can be interpreted as a transformation that re-configures the source space into a ``canonical form", i.e, a standardized representation that retains the original geometric properties. Additionally, Theorem \ref{thm: unviversal approximation} shows that if the true map $\phi$ is not available, we can approximate it by the neural network $\phi_{\omega}$ and still retain the overall architecture's universal approximation properties.

Note that, approximating $\widetilde{T}^* \circ \phi$ through $\widetilde{T}_{\theta} \circ \phi_{\omega}$, essentially entails the individual approximation of two GM-optimal maps $\widetilde{T}^*$ and $\phi$. Recall, that an isomorphism $\phi \in \Phi(\mu,\rho)$ is, itself, a solution to a GM problem, i.e $\text{GM}(\mu,\rho)$. Thus, we only need to define a \text{single} learning procedure for learning GM-optimal maps, which can be used for both $\widetilde{T}_{\theta}$ and $\phi_{\omega}$, adapted to their respective domains.

\subsection{Learning GM-optimal maps}\label{sec: learning gm optimal maps}


In this section, we start by introducing a general learning framework to approximate \textit{any} GM-optimal map between two arbitrary mm-spaces. Subsequently, based on the aforementioned framework, we introduce our learning procedure for training networks $\widetilde{T}_{\theta}$ and $\phi_{\omega}$. We base our method on the work by \citet{gap_monge}, who propose a loss for approximating OT maps between comparable spaces, i.e, Monge maps. Here, we briefly summarize their contribution and how our work is an extension of their framework to the GM case.

\textbf{The Monge Gap regularizer \citep{gap_monge}.}  Note that, the Monge problem between a source $\mu \in \mathcal{P}({\Omega})$ and target $\nu \in \mathcal{P}({\Omega})$ space, supported on the same domain ${\Omega}$, is defined as \citep{monge1781memoire}: 

\begin{equation}\label{eq: monge problem}
    M_{c}(\mu,\nu) = \inf_{T \# \mu=\nu} \int_{{\Omega}} c(\mathbf{x},T(\mathbf{x})) d\mu
\end{equation}

where $c: \Omega \times \Omega \longrightarrow \mathbb{R}_{+}$ is an intra-domain cost. 
Given any map $T$ which is used to approximate a solution to \eqref{eq: monge problem}, \citet{gap_monge} propose a loss function which quantifies its deviation from Monge optimality. Their loss consists of two terms: a \textit{fitting loss} and a \textit{regularizer} referred to as the \textit{Monge Gap}. The fitting loss quantifies how well $T$ satisfies the constraint $T \sharp \mu = \nu$. Conversely, the Monge gap  is defined as \citep{gap_monge}:
\begin{equation}\label{eq: monge-gap}
    \mathcal{M}(\mu,T \sharp \mu) = \int_{{\Omega}} c(\mathbf{x},T(\mathbf{x})) d\mu - M_{c}(\mu,T \# \mu)
\end{equation}

In essence, \eqref{eq: monge-gap} quantifies the deviation of T from being the c-optimal map between $\mu$ and $T \sharp \mu$, i.e, the map which induces the least amount of $c$ cost, while performing the push-forward operation $T \sharp \mu$. Naturally, when $T \sharp \mu = \nu$ is satisfied, i.e, when the fitting loss is zero, \eqref{eq: monge-gap} will quantify the deviation of $T$ from the solution of \eqref{eq: monge problem}.

\textbf{The Gromov-Monge Gap regularizer.}
We can extend the above elegant framework to the Gromov-Monge problem. For a given map $T: \mathcal{X} \longrightarrow \mathcal{Y}$, we define its induced \textit{Gromov-Monge (GM) gap} as follows:
\begin{equation}\label{eq: gm-gap}
      \mathcal{GM}^p(\mu,T \sharp \mu) = \text{dis}_p(T)^p
    -\text{GM}_{p}^p(\mu,T \sharp \mu)
\end{equation}

where 
$\text{GM}_{p}^p(\mu,T \sharp \mu)$ is the p-GM distance between measures $\mu$ and $T \sharp \mu$. Note that \eqref{eq: gm-gap} is a direct extension of \eqref{eq: monge-gap} to incomparable spaces. Intuitively, the first term represents the distortion induced by $T$
when moving mass from $\mu$ to $T \sharp \mu$. The second term represents the distortion induced by the GM-optimal map between the measures $\mu$ and $T \sharp \mu$. Thus, as in the Monge gap case, when $\mathcal{GM}^p(\mu,T \sharp \mu) = 0$ and $T \sharp \mu = \nu$, $T$ is theoretically guaranteed to be the solution $T^*$ to \eqref{eq: gromov-monge mu nu}.
Consequently, we can utilise the GM gap of \eqref{eq: gm-gap} as a regularizer in a loss function, designed to recover GM-optimal maps.

\textbf{GM-optimality loss.} Let ${T}_{\theta}$ be any parameterized map. 
Similar to the Monge-optimality loss proposed in \citet{gap_monge}, we propose the GM-optimality loss defined as:
\begin{equation}\label{eq: loss mu nu}
    \mathcal{L}(\theta) := \Delta({T}_{\theta} \sharp \mu, \nu) + \lambda \; \mathcal{GM}^p(\mu,{T}_{\theta} \sharp \mu) 
\end{equation}

where $\Delta({T}_{\theta} \sharp \mu, \nu)$ is the fitting loss and $\lambda$ is a regularization weight to stabilize training. 
We can use any valid discrepancy between measures on comparable spaces, such as the Sinkhorn divergence \citep{genevay2019sample}, i.e, 
$S_{c,\varepsilon}(\mu,T \sharp \mu)$
or the entropic Wasserstein distance, i.e, $W_{c,\varepsilon}(\mu,T \sharp \mu)$. Naturally, we get $\mathcal{L}(\theta) = 0$ when both terms of \eqref{eq: loss mu nu} are 0, i.e, when both probability mass and inter-domain distances are preserved. As such, for any family of parameterized maps $\{ T_\theta \}_\theta$ between two mm-spaces, we can recover the GM-optimal map by solving the optimization problem $T_{\theta} := \argmin_{\theta} \mathcal{L}(\theta)$.



\textbf{Optimization procedure.} Following the tripod structure of Figure \ref{fig: tripod}, the networks $\phi_\omega$ and $\widetilde{T}_\theta$ should be trained such that:
\begin{enumerate}
    \item $\phi_\omega$ moves mass from $\mu$ to $\rho$
in a GM-optimal way.
     \item $\widetilde{T}_\theta$ pushes forward the measure transformed by $\phi_{\omega}$, i.e, ${\rho}' := \phi_{\omega} \sharp \mu$, onto $\nu$ in a GM-optimal way.
\end{enumerate} 



Based on the GM-optimality loss of \eqref{eq: loss mu nu}, condition 1. can be formulated as the following optimization problem:
\begin{equation}\label{eq: phi omega optimization}
    \phi_{\omega} := \argmin_{\omega} \Delta({\phi}_{\omega} \sharp \mu, \rho) + \lambda \; \mathcal{GM}^2(\mu,{\phi}_{\omega} \sharp \mu)
\end{equation}
where $\mathcal{GM}^2(\mu,{\phi}_{\omega} \sharp \mu) = \text{dis}_2(\phi_{\omega})^2
    -\text{GM}_{2}^2(\mu,{\phi}_{\omega} \sharp \mu)$,
with $\text{dis}_2(\phi_{\omega})^2 := \lVert c_{\mathcal{X}} - c_{\mathcal{Z}} \lVert_{L^2(\mu_{\phi_{\omega}} \otimes \mu_{\phi_{\omega}})}$. Analogously, condition 2. can be formulated as:
\begin{equation}\label{eq: t theta optimization}
    \widetilde{T}_\theta := \argmin_{\theta} \Delta({\widetilde{T}}_{\theta} \sharp {\rho}', \nu) + \lambda \; \mathcal{GM}^2({\rho}',{\widetilde{T}}_{\theta} \sharp {\rho}')
\end{equation}
where $\mathcal{GM}^2(\rho',{\widetilde{T}}_{\theta} \sharp {\rho}')= \text{dis}_2(\widetilde{T}_{\theta})^2
    -\text{GM}^2_{2}({\rho}',{\widetilde{T}}_{\theta} \sharp {\rho}')$,
with $\text{dis}_2(\widetilde{T}_{\theta})^2 := \lVert c_{\mathcal{Z}} - c_{\mathcal{Y}} \lVert_{L^2({\rho'_{\widetilde{T}_{\theta}}} \otimes {\rho'_{\widetilde{T}_{\theta}}})}$. Given the theoretical properties of the GM-optimality loss and GM-gap, achieving zero in the objective functions of \eqref{eq: phi omega optimization} and \eqref{eq: t theta optimization} theoretically ensures that their solutions, $\phi_{\omega}$ and $\widetilde{T}_\theta$, are the GM-optimal maps $\phi$ and $\widetilde{T}^*$, respectively. Consequently, based on Theorem \ref{thm: unviversal approximation} and Proposition \ref{prop: gm_mn_rn_equality}, it will hold that $\widetilde{T}_\theta \circ \phi_\omega \approx \widetilde{T}^* \circ \phi$, thus approximating a GM-optimal map between $\mathcal{X}_\mu$  and $\mathcal{Y}_\nu$. We present the detailed learning procedure in Algorithm \ref{algo: composition}.
\begin{figure*}[ht]
\centering
\begin{minipage}{0.45\textwidth}
  \centering
  \begin{minipage}[b]{0.1\textwidth}
    \subcaption*{(a)}
  \end{minipage}
  \begin{minipage}[b]{0.8\textwidth}
    \includegraphics[width=\textwidth]{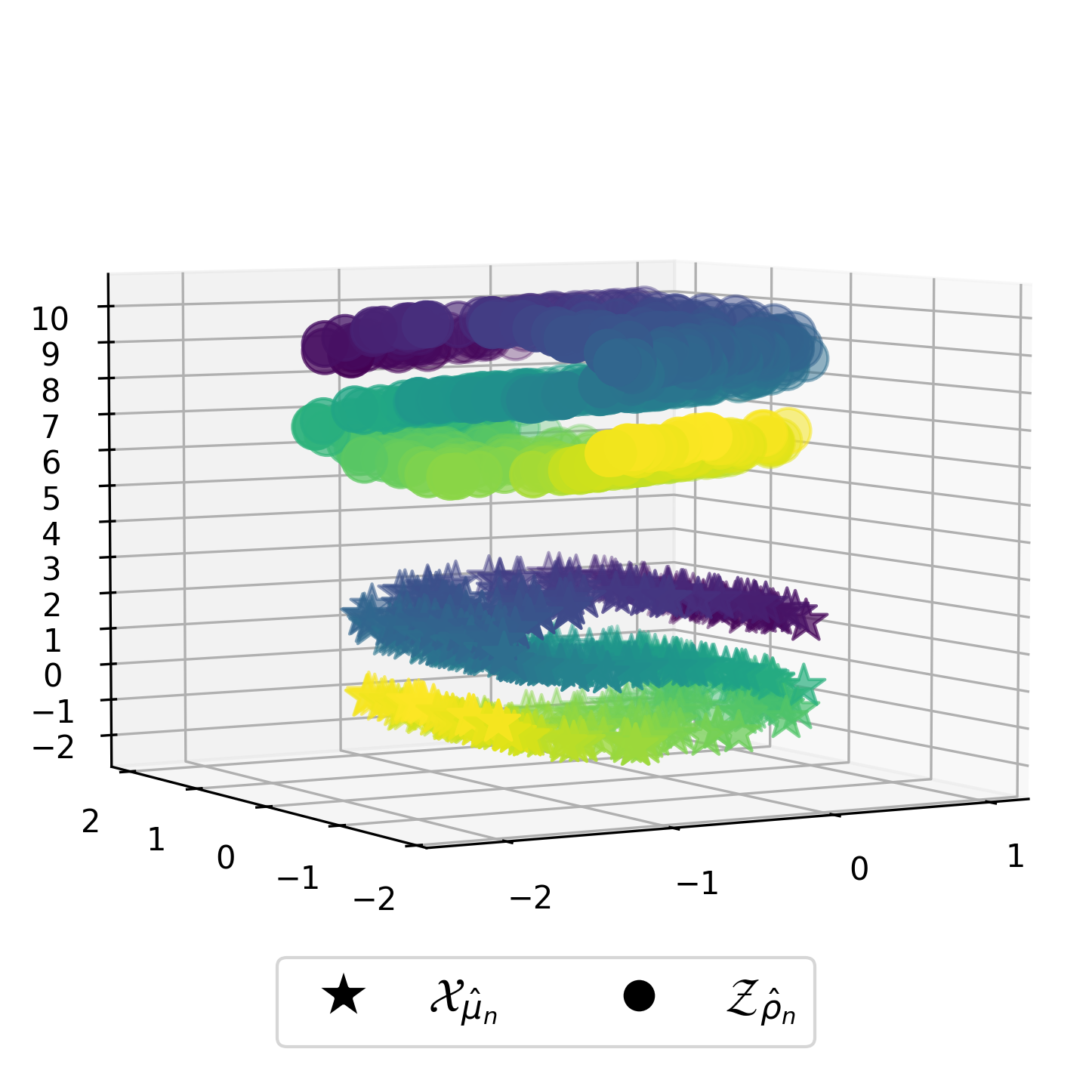}
  \end{minipage}
\end{minipage}
\hfill
\begin{minipage}{0.45\textwidth}
  \centering
  \begin{minipage}[b]{0.1\textwidth}
    \subcaption*{(b)}
  \end{minipage}
  \begin{minipage}[b]{0.8\textwidth}
    \includegraphics[width=\textwidth]{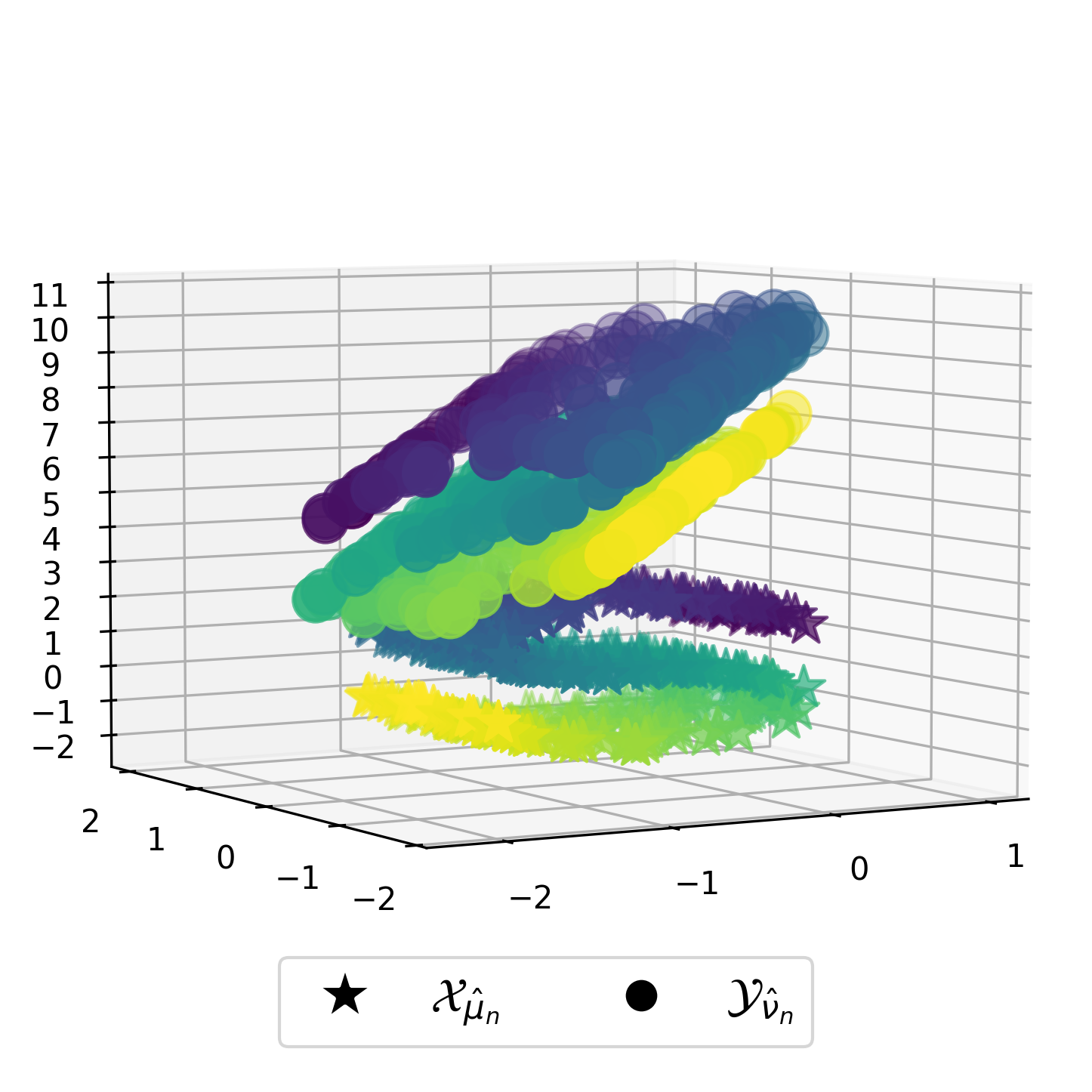}
  \end{minipage}
\end{minipage}
\caption{(a) \textbf{Isomorphic transformation of the source space.} 
Depiction of the empirical source and reference spaces $\mathbf{X}:=\mathcal{X}_{\hat{\mu}_n}$  and $\mathbf{Z}:=\mathcal{Z}_{\hat{\rho}_n}$ respectively. We obtain $\mathbf{Z}$ through the rigid transformation of $\mathbf{X}$, such that $\mathcal{X}_{\hat{\mu}_n} \cong^s \mathcal{Z}_{\hat{\rho}_n}$. To this end, we have $\mathbf{Z} = \phi(\mathbf{X}) := \mathbf{R}\mathbf{X} + \mathbf{t}$, where $\mathbf{R} \in \{ \mathbf{R} \in \mathbb{R}^{3 \times 3} \mid \mathbf{R}^T\mathbf{R} = \mathbf{R}\mathbf{R}^T = \mathbf{I} \}$ is an orthogonal (rotation) matrix and $\mathbf{t} \in \mathbb{R}^{3}$ a translation vector. (b) \textbf{Non-Isomorphic transformation of the reference space.}
Depiction of the empirical source and target spaces \(\mathbf{X} := \mathcal{X}_{\hat{\mu}_n}\) and \(\mathbf{Y} := \mathcal{Y}_{\hat{\nu}_n}\), respectively. The target point cloud \(\mathbf{Y}\) is obtained through a non-rigid transformation of the reference space $\mathbf{Z}$, i.e., $\mathbf{Y} = \widetilde{T}^*(\mathbf{Z}) = \mathbf{A}\mathbf{Z}$, where $\mathbf{A} \in \mathbb{R}^{3 \times 3}$ is a shearing matrix. Unlike $\phi$, the transport map $\widetilde{T}^*$ introduces a distortion of the spacial configuration of its input.}
\label{fig: transforms}
\end{figure*}
\begin{figure*}[ht]
\centering
\begin{minipage}{0.24\textwidth}
  \centering
   \begin{minipage}[b]{0.1\textwidth}
    \subcaption*{(a)}
  \end{minipage}
  \includegraphics[width=\textwidth]{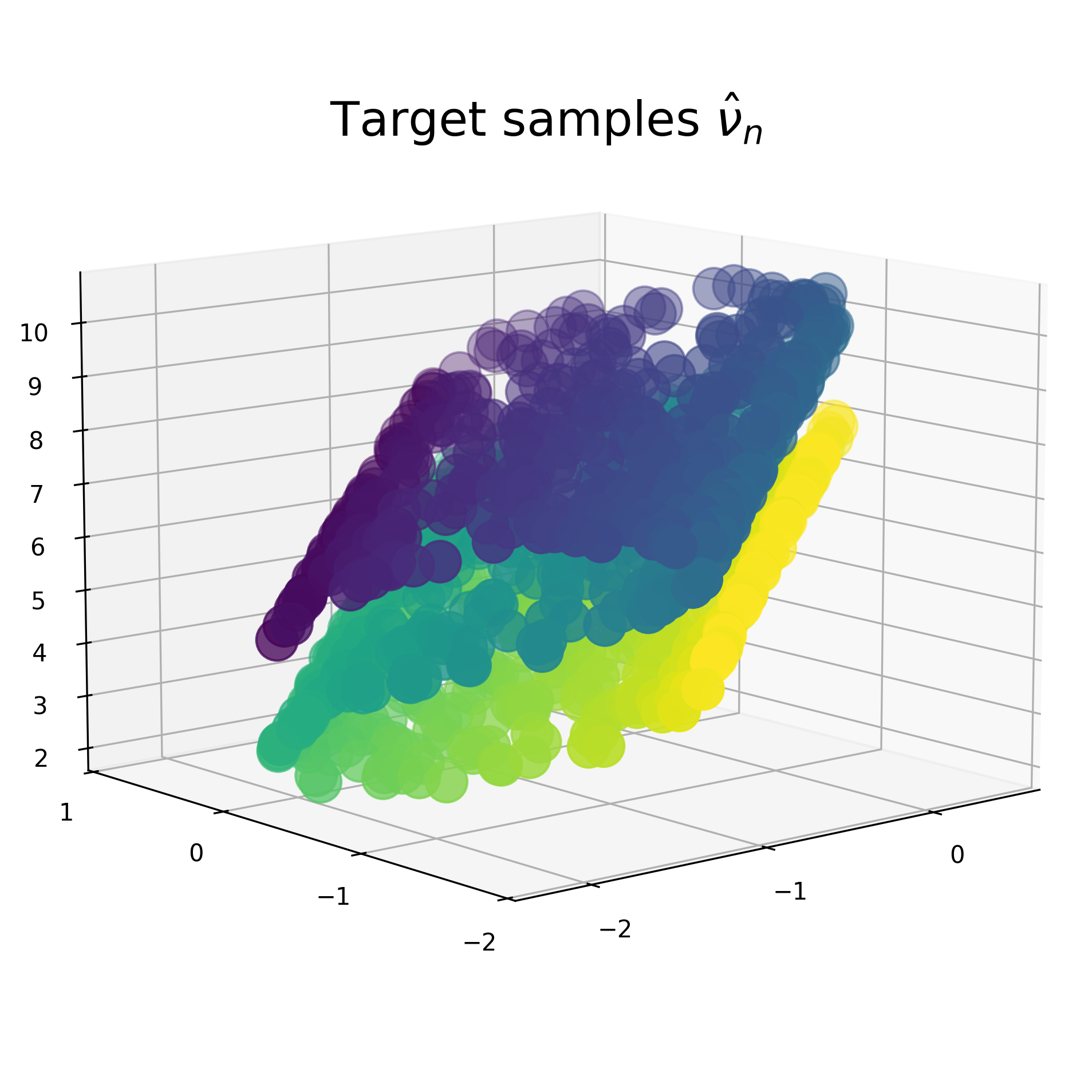}
\end{minipage}
\begin{minipage}{0.24\textwidth}
  \centering
  \begin{minipage}[b]{0.1\textwidth}
    \subcaption*{(b)}
    \end{minipage}
  \includegraphics[width=\textwidth]{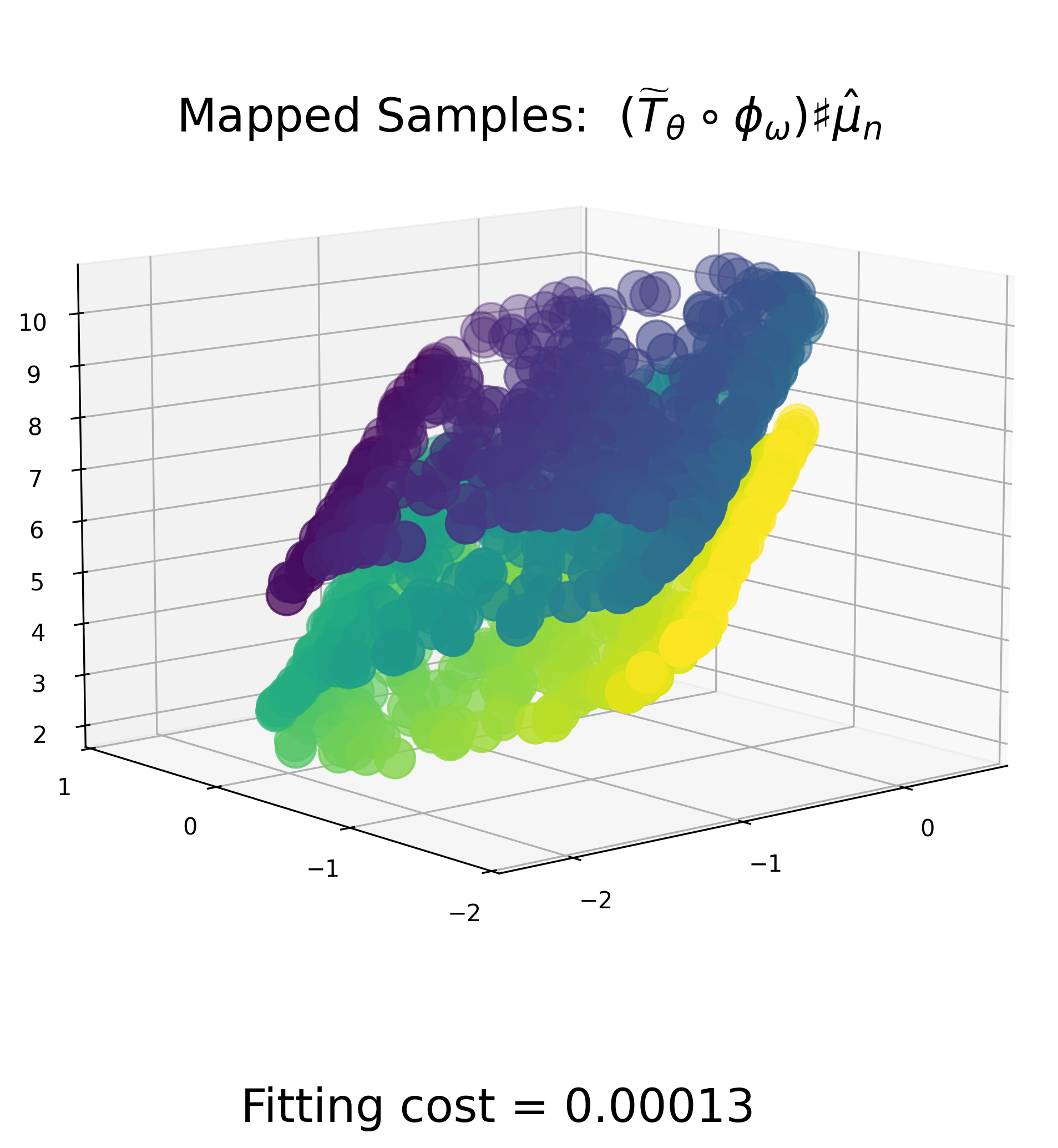}
\end{minipage}
\begin{minipage}{0.24\textwidth}
  \centering
  \begin{minipage}[b]{0.1\textwidth}
    \subcaption*{(c)}
    \end{minipage}
  \includegraphics[width=\textwidth]{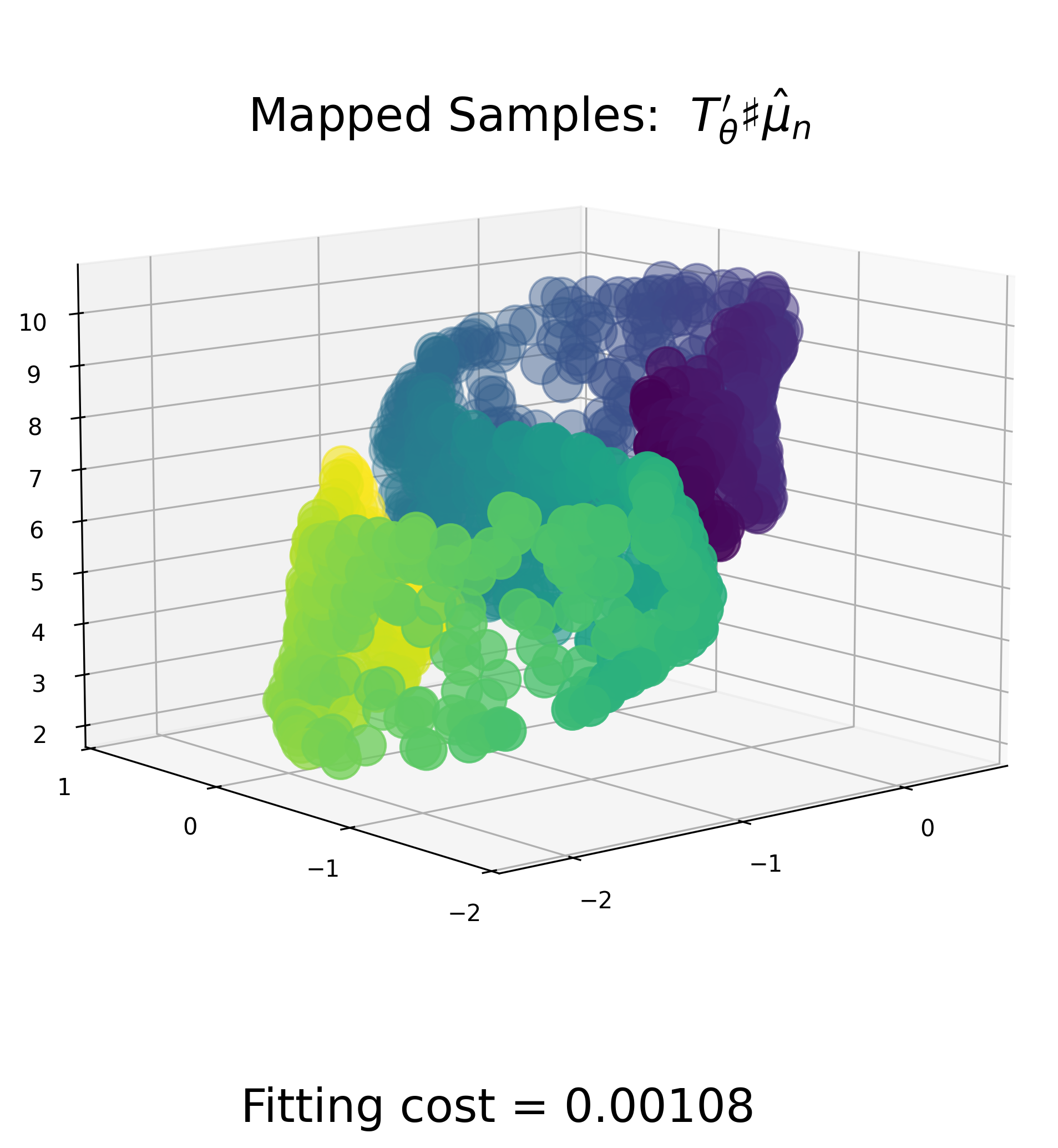}
\end{minipage}
\caption{ (a) \textbf{Ground truth target samples.} (b) \textbf{Mapped samples using neural composition.} Fitted samples using the learned composition map, i.e, $(\widetilde{T}_\theta \circ \phi_\omega) \sharp \hat{\mu}_n$. We first train $\phi_\omega$ for 5,000 iterations as a direct map between the source and reference samples. Then, using the pre-trained $\phi$ as initialization, we train the composition map $\widetilde{T}_\theta \circ \phi_\omega$, following Algorithm \ref{algo: composition} for $K_{outer} = 5$ and $K_{inner} = 2,000$ iterations. (c) \textbf{Mapped samples using direct parameterization.} Fitted samples using a direct map from the source to the target space $T_{\theta}' \sharp \hat{\mu}_n$ trained for 5,000 iterations. Unlike the direct map, the composition map clearly resembles the target distribution. We use the squared Euclidean Sinkhorn divergence, i.e, $S_{l_2^2,\varepsilon}$, with entropic regularization strength $\varepsilon = 0.1$ as the quantitative evaluation metric (i.e, evaluation fitting cost). We provide all details of our experimental framework in Appendix \ref{app: experiments}.}
\label{fig: results}
\end{figure*}

\textbf{Estimation from samples.} In practice,  
we have access to finite sets of samples, assumed to be drawn iid from the underlying, true, continuous distributions. As such, we consider the following sets consisting of $n$ number of samples: $\{ \mathbf{x}_i \}_{i=1}^{n}$, $\mathbf{x}_i \overset{\mathrm{iid}}{\sim} \mu$, $\{ \mathbf{y}_i \}_{i=1}^{n}$, $\mathbf{y}_i \overset{\mathrm{iid}}{\sim} \nu$ and $\{ \mathbf{z}_i \}_{i=1}^{n}$, $\mathbf{z}_i \overset{\mathrm{iid}}{\sim} \rho$. For the corresponding push-forward measures we have: 
$\{ \phi_\omega(\mathbf{x}_i) \}_{i=1}^{n}$, $\phi_\omega(\mathbf{x}_i) \overset{\mathrm{iid}}{\sim} \phi_{\omega} \sharp \mu \in \mathcal{P}(\mathcal{Z})$ and $\{ \widetilde{T}_\theta(\mathbf{z}'_i) \}_{i=1}^{n}$, $\widetilde{T}_\theta(\mathbf{z}'_i) \overset{\mathrm{iid}}{\sim} \widetilde{T}_\theta \sharp \rho' \in \mathcal{P}(\mathcal{Y})$, where $\mathbf{z}'_i = \phi_\omega(\mathbf{x}_i) \overset{\mathrm{iid}}{\sim} \phi_\omega \sharp \mu = \rho'$. The discrete empirical estimates of each corresponding (continuous) measure will, therefore, be: $\hat{\mu}_{n} = \frac{1}{n} \sum_{i=1}^{n} \delta_{\mathbf{x}_i}$, $\hat{\nu}_{n} = \frac{1}{n} \sum_{i=1}^{n} \delta_{\mathbf{y}_i}$, $\hat{\rho}_{n} = \frac{1}{n} \sum_{i=1}^{n} \delta_{\mathbf{z}_i}$, $\phi_\omega \sharp \hat{\mu}_n = \frac{1}{n} \sum_{i=1}^{n} \delta_{\phi_\omega(\mathbf{x}_i)}=\hat{\rho}'_n$
and $ \widetilde{T}_\theta \sharp \hat{\rho}'_n = \frac{1}{n} \sum_{i=1}^{n} \delta_{\widetilde{T}_\theta(\mathbf{z}'_i)}$. Given the above measure estimates, we can re-write each term in \eqref{eq: phi omega optimization} and \eqref{eq: t theta optimization} in a discrete formulation. Specifically, we can define the empirical estimate of the GM-gap $\mathcal{GM}^2(\mu,\phi_\omega \sharp \mu)$ as:

\begin{flalign}
\label{eq: gm-gap entropic-phi}
&\mathcal{GM}^2(\hat{\mu}_n,\phi_\omega \sharp \hat{\mu}_n) =   \hat{\text{dis}}_2(\phi_\omega){^2}-\textup{GW}_{2}^2(\hat{\mu}_n,\phi_\omega \sharp \hat{\mu}_n) &
\end{flalign}

where, \\
$\hat{\text{dis}}_2(\phi_\omega){^2} = \frac{1}{n^2} \sum_{i,j}^n   \left( c_{\mathcal{X}}(\mathbf{x}_i,\mathbf{x}_j)-c_{\mathcal{Z}}(\phi_\omega(\mathbf{x}_i),\phi_\omega(\mathbf{x}_j)) \right)^2$.
Analogously, for $\mathcal{GM}^2(\rho',\widetilde{T}_\theta \sharp \rho')$ we have:

\begin{flalign}
\label{eq: gm-gap entropic-T}
&\mathcal{GM}^2(\hat{\rho}'_n,\widetilde{T}_\theta \sharp \hat{\rho}'_n) =  \hat{\text{dis}}_2(\widetilde{T}_\theta){^2} -\textup{GW}_{2}^2(\hat{\rho}'_n,\widetilde{T}_\theta \sharp \hat{\rho}'_n) &
\end{flalign}

where, \\
$\hat{\text{dis}}_2(\widetilde{T}_\theta){^2} = \frac{1}{n^2} \sum_{i,j}^n \left( c_{\mathcal{Z}}(\mathbf{z}_i,\mathbf{z}_j)-c_{\mathcal{Y}}(\widetilde{T}_\theta(\mathbf{z}_i),\widetilde{T}_\theta(\mathbf{z}_j)) \right)^2$.

In both cases we use the GW distance with entropic regularization \citep{peyre2016,solomon2016} to approximate the GM distances $\text{GM}_{2}^2(\mu,{\phi}_{\omega} \sharp \mu)$ and $\text{GM}^2_{2}({\rho}',{\widetilde{T}}_{\theta} \sharp {\rho}')$ respectively. 
Note that according to Theorem 2 in \citet{mémoli2022comparison}, the GW distance is equivalent to the GM only for \textit{non-atomic} measures. Here we make the assumption that given enough samples $n$, the discrete entropic GW will approximate its continuous non-atomic counterpart and subsequently the corresponding GM distance. 

\begin{algorithm}
\caption{GM composition map estimation}\label{algo: composition}
\begin{algorithmic}[1] 
\STATE \textbf{Data:} Source $\mu$, target $\nu$ and reference $\rho$ measures accessible through empirical estimates $\hat{\mu}_n$, $\hat{\rho}_n$ and $\hat{\nu}_n$;
transport network $\widetilde{T}_{\theta}$;
isomorphism network $\phi_{\omega}$;
cost functions $c_{\mathcal{X}}$, $c_{\mathcal{Y}}$, and $c_{\mathcal{Z}}$;
regularization weight $\lambda$;
entropic regularization parameter $\varepsilon$; learning rate $\eta$; batch size $n$, number of iterations $K_{\text{outer}}$ and $K_{\text{inner}}$.
\STATE \textbf{Output:} Estimated GM-optimal map ${T} := \widetilde{T}_{\theta} \circ \phi_{\omega}$.

\FOR{$k = 1$ , \ldots, $K_{\text{outer}}$}
    \STATE Sample batches $\hat{\mu}_n$, $\hat{\rho}_n$
    \STATE Compute fitting loss $\Delta(\phi_{\omega} \sharp \hat{\mu}_n, \hat{\rho}_n)$
    \STATE Compute regularizer $\mathcal{GM}^2(\hat{\mu}_n, \phi_{\omega} \sharp \hat{\mu}_n)$
    \STATE $\hat{L}_{\phi}(\omega) \leftarrow \Delta(\phi_{\omega} \sharp \hat{\mu}_n, \hat{\rho}_n) + \lambda \; \mathcal{GM}^2(\hat{\mu}_n, \phi_{\omega} \sharp \hat{\mu}_n)$
    \STATE Update $\omega$ using $\frac{\partial \hat{L}_{\phi}(\omega)}{\partial \omega}$
    \FOR{$k = 1, \ldots, K_{\text{inner}}$}
        \STATE Re-sample batch $\hat{\mu}_n$
        \STATE Set $\hat{\rho}_n' \leftarrow \phi_{\omega} \sharp \hat{\mu}_n$
        \STATE Sample batch $\hat{\nu}_n$
        \STATE Compute fitting loss $\Delta(\widetilde{T}_{\theta} \sharp \hat{\rho}_n', \hat{\nu}_n)$
        \STATE Compute regularizer $\mathcal{GM}^2(\hat{\rho}_n', \widetilde{T}_{\theta} \sharp \hat{\rho}_n')$
        \STATE $\hat{L}_{T}(\theta) \leftarrow \Delta(\widetilde{T}_{\theta} \sharp \hat{\rho}_n', \hat{\nu}_n) + \lambda \; \mathcal{GM}^2(\hat{\rho}_n', \widetilde{T}_{\theta} \sharp \hat{\rho}_n')$
        \STATE Update $\theta$ using $\frac{\partial \hat{L}_{T}(\theta)}{\partial \theta}$
    \ENDFOR
\ENDFOR
\end{algorithmic}
\end{algorithm}

\section{Experiments}
In this section, we evaluate the ability of our method to recover GM-optimal maps on synthetic data. We implement our framework using the OTT-JAX\footnote{\href{https://github.com/ott-jax/ott}{https://github.com/ott-jax/ott}} package \citep{ott-jax}.

\textbf{Experimental tripod structure.}
 We consider a controlled setting, where the GM-optimal maps $\phi$ and $\widetilde{T}^*$ are known. We generate samples
 $\{ \mathbf{x}_i \}_{i=1}^{n}$ in $\mathbb{R}^3$, following the associated empirical measure $\hat{\mu}_n$ as described in section \ref{sec: learning gm optimal maps}. As such, we define the \textit{empirical} source mm-space $\mathcal{X}_{\hat{\mu}_n}$ as the 3D point cloud $\mathbf{X} := \mathcal{X}_{\hat{\mu}_n} \in \mathbb{R}^{n \times 3}$. Following the tripod structure of Figure \ref{fig: tripod}, we apply a known isomorphism, i.e a \textit{rigid} transformation $\phi$ to $\mathcal{X}_{\hat{\mu}_n}$, to obtain the (empirical) reference mm-space, i.e, $\mathbf{Z} = \phi(\mathbf{X}) := \mathcal{Z}_{\hat{\rho}_n} \in \mathbb{R}^{n \times 3}$ (Figure \ref{fig: transforms} (a)). Subsequently, we generate the empirical target space by applying a \textit{non-rigid} transformation, $\widetilde{T}^*$, to the reference point-cloud $\mathbf{Z}$, i.e $\mathbf{Y} = \widetilde{T}^*(\mathbf{Z}) := \mathcal{Y}_{\hat{\nu}_n} \in \mathbb{R}^{n \times 3}$ (Figure \ref{fig: transforms} (b)). 

 \textbf{Results.} Following Algorithm \ref{algo: composition}, we train neural networks $\phi_{\omega}$ and $\widetilde{T}_{\theta}$
 on the empirical sample sets. Since the target point cloud $\mathbf{Y}$ is the result of the GM-optimal transformation $\widetilde{T}^* \circ \phi$, if $\widetilde{T}_{\theta} \circ \phi_{\omega}$ approximates GM-optimality, we expect the mapped points to approximate $\mathbf{Y}$, i.e $\widetilde{T}_{\theta} \circ \phi_{\omega}(\mathbf{X}) \approx \mathbf{Y}$. We present the mapped samples of the learned composition in Figure \ref{fig: results} (b), and compare them to the ground truth target samples (Figure \ref{fig: results} (a)). Indeed, the learned map $\widetilde{T}_{\theta} \circ \phi_{\omega}$ is able to fully match the ground truth target point cloud. 
 To validate the importance of learning a  composition instead of a direct approximation, we present the equivalent results when learning a direct mapping $T_{\theta}': \mathcal{X}_{\hat{\mu}_n} \longrightarrow \mathcal{Y}_{\hat{\nu}_n} $, from the source to the target point cloud (Figure \ref{fig: results} (c)). In this case, we train $T_{\theta}': \mathcal{X}_{\hat{\mu}_n}$ on a single loop using the loss $\hat{L}_{T'}(\theta) \leftarrow \Delta(T_{\theta}'\sharp \hat{\mu}_{n},\hat{\nu}_{n})  + \lambda \; \mathcal{GM}^2(\hat{\mu}_n,T_{\theta}' \sharp \hat{\mu}_n)$. Unlike the composition map, the direct mapping is unable to accurately match the geometry of the target distribution. 

\section{Discussion}
In this work, we introduce a theoretically grounded framework to approximate optimal transport maps between incomparable spaces, based on the geometric property of invariance to strong isomorphism. We show that the GM-optimal map between a source and target distribution can be decomposed into an isomorphism and a subsequent GM-optimal map, which can be adequately approximated by two separate neural networks. Intuitively, this implies the hypothesis that any geometrically informed transformation between two distributions is a sequence of a rigid and a non-rigid transformation. This suggests that by parameterizing (and training) each transformation step independently, we more effectively approximate the true optimal transportation map between incomparable distributions. Empirical results on toy data reinforce this hypothesis. We believe this could open a new avenue to address the fundamental problem of geometrically informed transformations of samples between distributions.

\nocite{langley00}

\bibliography{example_paper}

\begin{thebibliography}{42}
\providecommand{\natexlab}[1]{#1}
\providecommand{\url}[1]{\texttt{#1}}
\expandafter\ifx\csname urlstyle\endcsname\relax
  \providecommand{\doi}[1]{doi: #1}\else
  \providecommand{\doi}{doi: \begingroup \urlstyle{rm}\Url}\fi

\bibitem[Alvarez-Melis \& Fusi(2020)Alvarez-Melis and Fusi]{alvarez2020geometric}
Alvarez-Melis, D. and Fusi, N.
\newblock Geometric dataset distances via optimal transport.
\newblock \emph{Advances in Neural Information Processing Systems}, 33:\penalty0 21428--21439, 2020.

\bibitem[Alvarez-Melis et~al.(2019)Alvarez-Melis, Jegelka, and Jaakkola]{alvarez2019towards}
Alvarez-Melis, D., Jegelka, S., and Jaakkola, T.~S.
\newblock Towards optimal transport with global invariances.
\newblock In \emph{The 22nd International Conference on Artificial Intelligence and Statistics}, pp.\  1870--1879. PMLR, 2019.

\bibitem[Arjovsky et~al.(2017)Arjovsky, Chintala, and Bottou]{arjovsky2017wasserstein}
Arjovsky, M., Chintala, S., and Bottou, L.
\newblock Wasserstein generative adversarial networks.
\newblock In \emph{International conference on machine learning}, pp.\  214--223. PMLR, 2017.

\bibitem[Brenier(1987)]{brenier1987decomposition}
Brenier, Y.
\newblock D{\'e}composition polaire et r{\'e}arrangement monotone des champs de vecteurs.
\newblock \emph{CR Acad. Sci. Paris S{\'e}r. I Math.}, 305:\penalty0 805--808, 1987.

\bibitem[Bunne et~al.(2019)Bunne, Alvarez-Melis, Krause, and Jegelka]{pmlr-v97-bunne19a}
Bunne, C., Alvarez-Melis, D., Krause, A., and Jegelka, S.
\newblock Learning generative models across incomparable spaces.
\newblock In Chaudhuri, K. and Salakhutdinov, R. (eds.), \emph{Proceedings of the 36th International Conference on Machine Learning}, volume~97 of \emph{Proceedings of Machine Learning Research}, pp.\  851--861. PMLR, 09--15 Jun 2019.
\newblock URL \url{https://proceedings.mlr.press/v97/bunne19a.html}.

\bibitem[Bunne et~al.(2023)Bunne, Stark, Gut, Del~Castillo, Levesque, Lehmann, Pelkmans, Krause, and R{\"a}tsch]{bunne2023learning}
Bunne, C., Stark, S.~G., Gut, G., Del~Castillo, J.~S., Levesque, M., Lehmann, K.-V., Pelkmans, L., Krause, A., and R{\"a}tsch, G.
\newblock Learning single-cell perturbation responses using neural optimal transport.
\newblock \emph{Nature Methods}, 20\penalty0 (11):\penalty0 1759--1768, 2023.

\bibitem[{Chowdhury} \& {M{\'e}moli}(2018){Chowdhury} and {M{\'e}moli}]{memoli_2019}
{Chowdhury}, S. and {M{\'e}moli}, F.
\newblock {The Gromov-Wasserstein distance between networks and stable network invariants}.
\newblock \emph{arXiv e-prints}, art. arXiv:1808.04337, August 2018.
\newblock \doi{10.48550/arXiv.1808.04337}.

\bibitem[Courty et~al.(2017)Courty, Flamary, Habrard, and Rakotomamonjy]{courty2017joint}
Courty, N., Flamary, R., Habrard, A., and Rakotomamonjy, A.
\newblock Joint distribution optimal transportation for domain adaptation.
\newblock \emph{Advances in neural information processing systems}, 30, 2017.

\bibitem[Creswell et~al.(2018)Creswell, White, Dumoulin, Arulkumaran, Sengupta, and Bharath]{gans}
Creswell, A., White, T., Dumoulin, V., Arulkumaran, K., Sengupta, B., and Bharath, A.~A.
\newblock Generative adversarial networks: An overview.
\newblock \emph{IEEE signal processing magazine}, 35\penalty0 (1):\penalty0 53--65, 2018.

\bibitem[Cuturi(2013)]{cuturi2013sinkhorn}
Cuturi, M.
\newblock Sinkhorn distances: Lightspeed computation of optimal transportation distances, 2013.

\bibitem[Cuturi et~al.(2022)Cuturi, Meng-Papaxanthos, Tian, Bunne, Davis, and Teboul]{ott-jax}
Cuturi, M., Meng-Papaxanthos, L., Tian, Y., Bunne, C., Davis, G., and Teboul, O.
\newblock Optimal transport tools (ott): A jax toolbox for all things wasserstein.
\newblock \emph{arXiv preprint arXiv:2201.12324}, 2022.

\bibitem[Demetci et~al.(2022)Demetci, Santorella, Sandstede, Noble, and Singh]{demetci2022scot}
Demetci, P., Santorella, R., Sandstede, B., Noble, W.~S., and Singh, R.
\newblock Scot: single-cell multi-omics alignment with optimal transport.
\newblock \emph{Journal of computational biology}, 29\penalty0 (1):\penalty0 3--18, 2022.

\bibitem[Dumont et~al.(2024)Dumont, Lacombe, and Vialard]{dumont2024existence}
Dumont, T., Lacombe, T., and Vialard, F.-X.
\newblock On the existence of monge maps for the gromov--wasserstein problem.
\newblock \emph{Foundations of Computational Mathematics}, pp.\  1--48, 2024.

\bibitem[Fan et~al.(2023)Fan, Liu, Ma, Zhou, and Chen]{fan2023neural}
Fan, J., Liu, S., Ma, S., Zhou, H., and Chen, Y.
\newblock Neural monge map estimation and its applications.
\newblock \emph{Transactions on Machine Learning Research}, 2023.
\newblock ISSN 2835-8856.

\bibitem[Fickinger et~al.(2021)Fickinger, Cohen, Russell, and Amos]{fickinger2021cross}
Fickinger, A., Cohen, S., Russell, S., and Amos, B.
\newblock Cross-domain imitation learning via optimal transport.
\newblock \emph{arXiv preprint arXiv:2110.03684}, 2021.

\bibitem[Folland(1999)]{folland1999real}
Folland, G.~B.
\newblock \emph{Real analysis: modern techniques and their applications}, volume~40.
\newblock John Wiley \& Sons, 1999.

\bibitem[Genevay et~al.(2019)Genevay, Chizat, Bach, Cuturi, and Peyré]{genevay2019sample}
Genevay, A., Chizat, L., Bach, F., Cuturi, M., and Peyré, G.
\newblock Sample complexity of sinkhorn divergences, 2019.

\bibitem[Kantorovich(2006)]{kantorovich2006translocation}
Kantorovich, L.~V.
\newblock On the translocation of masses.
\newblock \emph{Journal of mathematical sciences}, 133\penalty0 (4):\penalty0 1381--1382, 2006.

\bibitem[Kingma \& Ba(2017)Kingma and Ba]{kingma2017adam}
Kingma, D.~P. and Ba, J.
\newblock Adam: A method for stochastic optimization, 2017.

\bibitem[Klein et~al.(2024)Klein, Uscidda, Theis, and Cuturi]{klein2024entropic}
Klein, D., Uscidda, T., Theis, F., and Cuturi, M.
\newblock Entropic (gromov) wasserstein flow matching with genot, 2024.

\bibitem[Korotin et~al.(2020)Korotin, Egiazarian, Asadulaev, Safin, and Burnaev]{korotin2020wasserstein2}
Korotin, A., Egiazarian, V., Asadulaev, A., Safin, A., and Burnaev, E.
\newblock Wasserstein-2 generative networks, 2020.

\bibitem[Korotin et~al.(2023)Korotin, Selikhanovych, and Burnaev]{korotin2023neural_OT}
Korotin, A., Selikhanovych, D., and Burnaev, E.
\newblock Neural optimal transport.
\newblock In \emph{The Eleventh International Conference on Learning Representations}, 2023.
\newblock URL \url{https://openreview.net/forum?id=d8CBRlWNkqH}.

\bibitem[Kratsios \& Bilokopytov(2020)Kratsios and Bilokopytov]{kratsios2020non}
Kratsios, A. and Bilokopytov, I.
\newblock Non-euclidean universal approximation.
\newblock \emph{Advances in Neural Information Processing Systems}, 33:\penalty0 10635--10646, 2020.

\bibitem[Langley(2000)]{langley00}
Langley, P.
\newblock Crafting papers on machine learning.
\newblock In Langley, P. (ed.), \emph{Proceedings of the 17th International Conference on Machine Learning (ICML 2000)}, pp.\  1207--1216, Stanford, CA, 2000. Morgan Kaufmann.

\bibitem[Makkuva et~al.(2020)Makkuva, Taghvaei, Oh, and Lee]{makkuva2020optimal}
Makkuva, A.~V., Taghvaei, A., Oh, S., and Lee, J.~D.
\newblock Optimal transport mapping via input convex neural networks, 2020.

\bibitem[M{\'e}moli \& Needham(2022)M{\'e}moli and Needham]{memoli2022distance}
M{\'e}moli, F. and Needham, T.
\newblock Distance distributions and inverse problems for metric measure spaces.
\newblock \emph{Studies in Applied Mathematics}, 149\penalty0 (4):\penalty0 943--1001, 2022.

\bibitem[Monge(1781)]{monge1781memoire}
Monge, G.
\newblock M{\'e}moire sur la th{\'e}orie des d{\'e}blais et des remblais.
\newblock \emph{Mem. Math. Phys. Acad. Royale Sci.}, pp.\  666--704, 1781.

\bibitem[Mémoli(2011)]{memoli_2011}
Mémoli, F.
\newblock Gromov-wasserstein distances and the metric approach to object matching.
\newblock \emph{Foundations of Computational Mathematics}, 11\penalty0 (4):\penalty0 417--487, 2011.
\newblock URL \url{http://dblp.uni-trier.de/db/journals/focm/focm11.html#Memoli11}.

\bibitem[Mémoli \& Needham(2022{\natexlab{a}})Mémoli and Needham]{mémoli2022comparison}
Mémoli, F. and Needham, T.
\newblock Comparison results for gromov-wasserstein and gromov-monge distances, 2022{\natexlab{a}}.

\bibitem[Mémoli \& Needham(2022{\natexlab{b}})Mémoli and Needham]{mémoli2022distancedistributionsinverseproblems}
Mémoli, F. and Needham, T.
\newblock Distance distributions and inverse problems for metric measure spaces, 2022{\natexlab{b}}.
\newblock URL \url{https://arxiv.org/abs/1810.09646}.

\bibitem[Nekrashevich et~al.(2023)Nekrashevich, Korotin, and Burnaev]{nekrashevich2023neural}
Nekrashevich, M., Korotin, A., and Burnaev, E.
\newblock Neural gromov-wasserstein optimal transport.
\newblock \emph{arXiv preprint arXiv:2303.05978}, 2023.

\bibitem[Peyr{\'e} et~al.(2016)Peyr{\'e}, Cuturi, and Solomon]{peyre2016}
Peyr{\'e}, G., Cuturi, M., and Solomon, J.
\newblock Gromov-wasserstein averaging of kernel and distance matrices.
\newblock In \emph{International conference on machine learning}, pp.\  2664--2672. PMLR, 2016.

\bibitem[Rezende \& Mohamed(2015)Rezende and Mohamed]{rezende2015variational}
Rezende, D. and Mohamed, S.
\newblock Variational inference with normalizing flows.
\newblock In \emph{International conference on machine learning}, pp.\  1530--1538. PMLR, 2015.

\bibitem[Rout et~al.(2022)Rout, Korotin, and Burnaev]{rout2022generative}
Rout, L., Korotin, A., and Burnaev, E.
\newblock Generative modeling with optimal transport maps, 2022.

\bibitem[Salmona et~al.(2021)Salmona, Delon, and Desolneux]{salmona2021gromov}
Salmona, A., Delon, J., and Desolneux, A.
\newblock Gromov-wasserstein distances between gaussian distributions.
\newblock \emph{arXiv preprint arXiv:2104.07970}, 2021.

\bibitem[Santambrogio(2015)]{santambrogio2015optimal}
Santambrogio, F.
\newblock \emph{Optimal Transport for Applied Mathematicians: Calculus of Variations, PDEs, and Modeling}.
\newblock Progress in Nonlinear Differential Equations and Their Applications. Springer International Publishing, 2015.
\newblock ISBN 9783319208282.
\newblock URL \url{https://books.google.ch/books?id=UOHHCgAAQBAJ}.

\bibitem[Sebbouh et~al.(2024)Sebbouh, Cuturi, and Peyr{\'e}]{sebbouh2024structured}
Sebbouh, O., Cuturi, M., and Peyr{\'e}, G.
\newblock Structured transforms across spaces with cost-regularized optimal transport.
\newblock In \emph{International Conference on Artificial Intelligence and Statistics}, pp.\  586--594. PMLR, 2024.

\bibitem[Solomon et~al.(2016)Solomon, Peyr{\'e}, Kim, and Sra]{solomon2016}
Solomon, J., Peyr{\'e}, G., Kim, V.~G., and Sra, S.
\newblock Entropic metric alignment for correspondence problems.
\newblock \emph{ACM Transactions on Graphics (ToG)}, 35\penalty0 (4):\penalty0 1--13, 2016.

\bibitem[Song et~al.(2020)Song, Sohl-Dickstein, Kingma, Kumar, Ermon, and Poole]{song2020score}
Song, Y., Sohl-Dickstein, J., Kingma, D.~P., Kumar, A., Ermon, S., and Poole, B.
\newblock Score-based generative modeling through stochastic differential equations.
\newblock \emph{arXiv preprint arXiv:2011.13456}, 2020.

\bibitem[Sturm(2020)]{sturm2012}
Sturm, K.-T.
\newblock The space of spaces: curvature bounds and gradient flows on the space of metric measure spaces, 2020.

\bibitem[Uscidda \& Cuturi(2023)Uscidda and Cuturi]{gap_monge}
Uscidda, T. and Cuturi, M.
\newblock The monge gap: A regularizer to learn all transport maps, 2023.

\bibitem[Vayer(2020)]{vayer_phd}
Vayer, T.
\newblock A contribution to optimal transport on incomparable spaces.
\newblock \emph{arXiv preprint arXiv:2011.04447}, 2020.

\end{thebibliography}
\bibliographystyle{icml2024}

\newpage
\appendix
\onecolumn

\nocite{langley00}



\newpage
\appendix
\onecolumn

\section{Proofs}
\subsection{Proof of Proposition \ref{prop: gm_mn_rn_equality}}\label{app: proof prop gm equal}

Since $\mathcal{X}_{\mu} \cong^{s} \mathcal{Z}_{\rho}$,  $\exists \; \phi \in \Phi(\mu,\rho)$, which is a bijection. Thus, $\exists \; \phi^{-1} \in \Phi(\rho,\mu)$, s.t $GM(\rho,\mu)=0$. Since GM defines a Lawvere metric on $\mathcal{M}_p$ (see \citet{mémoli2022distancedistributionsinverseproblems} and Theorem 1 in \citet{mémoli2022comparison}),
following the triangle inequality we have
$\text{GM}(\mu,\nu) \leq \text{GM}(\mu,\rho) + \text{GM}(\rho,\nu)$ and $\text{GM}(\rho,\nu) \leq \text{GM}(\rho,\mu) + \text{GM}(\mu,\nu)$. Combining the above inequalities and since $\text{GM}(\mu,\rho)=0$, we get $\text{GM}(\mu,\nu)=\text{GM}(\rho,\nu)$.

\subsection{Proof of Proposition \ref{prop: cgm=gm}}\label{app: cgm=gm}

Constraining the optimal transport plan to be within $\mathcal{I}(\mu,\nu)$, will result in a \textit{constraint} GM-problem:

\begin{align}\label{eq:gm-constraint-mn-app}
\begin{split}
\text{CGM}_p(\mu,\nu) = 
    \inf_{T \in \mathcal{I}(\mu,\nu)} \left( \iint_{\mathcal{X} \times \mathcal{X}} \lvert c_{\mathcal{X}}(\mathbf{x},\mathbf{x'})-c_{\mathcal{Y}}(T(\mathbf{x}),T(\mathbf{x'})) \lvert^p
    \,d\mu(\mathbf{x})  \,d\mu(\mathbf{x'}) \right)^{1/p} =\\
        \inf_{(\widetilde{T} \circ \phi) \# \mu = \nu } \left( \iint_{\mathcal{X} \times \mathcal{X}} \lvert c_{\mathcal{X}}(\mathbf{x},\mathbf{x'})-c_{\mathcal{Y}}((\widetilde{T} \circ \phi)(\mathbf{x}),(\widetilde{T} \circ \phi)(\mathbf{x'})) \lvert^p
    \,d\mu(\mathbf{x})  \,d\mu(\mathbf{x'}) \right)^{1/p}
\end{split}
\end{align}

Based on Definition \ref{def: strong-isomorphism}, any isomorphism $\phi \in {\Phi}(\mu,\rho)$ is an invertible bijective map. Thus, for every pair $(\mathbf{z},\mathbf{z'}) \in \mathcal{Z} \times \mathcal{Z}$ there exists a unique pair $(\mathbf{x},\mathbf{x'}) \in \mathcal{X} \times \mathcal{X}$ such that $\mathbf{z}=\phi(\mathbf{x})$ and $\mathbf{z'}=\phi(\mathbf{x'})$. Additionally, $\phi^{-1}: \mathcal{Z} \longrightarrow \mathcal{X}$ is also a bijective isometry. Consequently, for every $\mathbf{z} \in \mathcal{Z}$ there is a unique $\mathbf{x} \in \mathcal{X}$ s.t $\mathbf{x} = \phi^{-1}(\mathbf{z})$.  Also, since $\phi \# \mu = \rho$, it holds for every $B \subseteq \mathcal{Z}$ that $\rho(B) = \mu(\phi^{-1}(B))$.
Considering all the above, we are permitted to perform a change of variables over the integral in \eqref{eq:gm-constraint-mn-app}. Thus, we have:

\begin{align}\label{eq: gm-source-target-phi-change-of-vars}
\begin{split}
    \text{CGM}_{p}(\mu,\nu) = 
    \inf_{\widetilde{T}{\# \rho}=\nu} \left( \iint_{\mathcal{X} \times \mathcal{X}} \lvert c_{\mathcal{Z}}(\mathbf{\phi(x)},\mathbf{\phi(x')})-c_{\mathcal{Y}}((\widetilde{T} \circ \phi)(\mathbf{x}),(\widetilde{T} \circ \phi)(\mathbf{x'})) \lvert^p
    \,d\mu(\mathbf{x})  \,d\mu(\mathbf{x'}) \right)^{1/p} = \\
     \inf_{\widetilde{T}{\# \rho}=\nu} \left( \iint_{\mathcal{Z} \times \mathcal{Z}} \lvert c_{\mathcal{Z}}(\mathbf{z},\mathbf{z'})-c_{\mathcal{Y}}(\widetilde{T}(\mathbf{z}),\widetilde{T} (\mathbf{z'})) \lvert^p
    \,d\mu(\phi^{-1}(\mathbf{z}))  \,d\mu(\phi^{-1}(\mathbf{z'}))   \right)^{1/p} = \\
         \inf_{\widetilde{T}{\# \rho}=\nu} \left( \iint_{\mathcal{Z} \times \mathcal{Z}} \lvert c_{\mathcal{Z}}(\mathbf{z},\mathbf{z'})-c_{\mathcal{Y}}(\widetilde{T}(\mathbf{z}),\widetilde{T} (\mathbf{z'})) \lvert^p
    \,d\rho(\mathbf{z})  \,d\rho(\mathbf{z'})  \right)^{1/p} = \\
    \text{GM}_{p}(\rho,\nu)
\end{split}
\end{align}

Then, following Proposition \ref{prop: gm_mn_rn_equality} we get $\text{CGM}_{p}(\mu,\nu)= \text{GM}_{p}(\rho,\nu) = \text{GM}_{p}(\mu,\nu)$.

\subsection{Proof of Theorem \ref{thm: unviversal approximation}}\label{app: proof theorem universal}

Since $\mathcal{X}_{\mu} \cong^{s} \mathcal{Z}_{\rho}$, there exists a measure preserving isometry $\phi \in \Phi(\mu,\rho)$ between $\mathcal{X}$ and $\mathcal{Z}$. Since $\phi$ is an isometry it will be continuous and injective. According to \citet{kratsios2020non}, for any given continuous and injective function $\phi: \mathcal{X} \longrightarrow \mathcal{Z}$, the collection of functions 
$f \circ \phi \in C_c(\mathcal{X,\mathcal{Y}})$,
where $f: \mathcal{Z} \longrightarrow  \mathcal{Y}$ is a deep-forward ReLU neural network, will be dense in 
$C_c(\mathcal{X},\mathcal{Y})$. Following Proposition 7.9 in \citet{folland1999real},  $C_c(\mathcal{X},\mathcal{Y})$ is dense in $L^2_{\mu}(\mathcal{X},\mathcal{Y})$. Consequently, the set of functions $f \circ \phi$ is also dense in  
$L^2_{\mu}(\mathcal{X},\mathcal{Y})$. 
Following the Proof of Theorem 1 in \citet{korotin2023neural_OT}, it is straightforward to show that since $\nu$ has a finite second moment, for any transport $T \in \mathcal{T}(\mu,\nu)$ map between $\mu$ and $\nu$ 
we have $T \in L^2_{\mu}(\mathcal{X},\mathcal{Y})$.

Thus, there exists a ReLU neural network $f_{\theta}$ s.t for any $ \epsilon >0$ we have:

\begin{equation}\label{ineq T-fcircphi}
    \lVert T - f_{\theta} \circ \phi \lVert_{L^2_{\mu}} \leq \epsilon/2
\end{equation}

It holds that $f_{\theta} \circ \phi, f_{\theta} \circ \phi_{\omega} \in C_c(\mathcal{X},\mathcal{Y})$ and $\phi,\phi_{\omega} \in C_c(\mathcal{X},\mathcal{Z})$. Continuous functions with compact support are 
also p-integrable w.r.t any finite measure in their domain. Thus it holds: 
$f_{\theta} \circ \phi, f_{\theta} \circ \phi_{\omega} \in L^2_{\mu}(\mathcal{X},\mathcal{Y})$ and $\phi,\phi_{\omega} \in L^2_{\mu}(\mathcal{X},\mathcal{Z})$.

Therefore, the norms $\lVert T - f_{\theta} \circ \phi_{\omega} \lVert_{L^2_{\mu}}$, $\lVert f_{\theta} \circ \phi - f_{\theta} \circ \phi_{\omega} \lVert_{L^2_{\mu}}$ and $\lVert \phi - \phi_{\omega}\lVert_{L^2_{\mu}}$ are well-defined and for any $T$ and $\phi_{\omega}$ we have:

\begin{align}\label{ineq T-fcircphiomega}
\begin{split}
  \lVert T - f_{\theta} \circ \phi_{\omega} \lVert_{L^2_{\mu}} = \lVert T -f_{\theta} \circ \phi + f_{\theta} \circ \phi - f_{\theta} \circ \phi_{\omega}\lVert_{L^2_{\mu}} \\
\leq  \lVert T -f_{\theta} \circ \phi \lVert_{L^2_{\mu}} + \lVert  f_{\theta} \circ \phi - f_{\theta} \circ \phi_{\omega} \lVert_{L^2_{\mu}} 
\end{split}
\end{align}

where the inequality follows from Minkowski’s inequality.

We will now focus on deriving a bound for the term $\lVert  f_{\theta} \circ \phi - f_{\theta} \circ \phi_{\omega} \lVert_{L^2_{\mu}} $ in inequality \ref{ineq T-fcircphiomega}.

Every feed-forward ReLU neural network is Lipschitz continuous w.r.t the $L_2$ norm. Thus, $\exists$ a constant $L >0$ s.t  $\forall z,z' \in \mathcal{Z}$:

\begin{equation}\label{ineq: ftheta lipschitz}
    \lVert f_{\theta}(z) - f_{\theta}(z') \lVert_2 \leq L\lVert z - z' \lVert_2
\end{equation}

We have 
$\phi_{\omega} : \mathcal{X} \longrightarrow \mathcal{Z}$ 
and 
$\phi :  
\mathcal{X}  \longrightarrow \mathcal{Z}$
, being two distinct mappings with the same domain and codomain.

Thus, $\forall x \in \mathcal{X}$ we have $\phi(x),\phi_{\omega}(x) \in \mathcal{Z}$. Note that it doesn't necessarily hold that $\phi(x) = \phi_{\omega}(x)$. As such,
we can assign $z=\phi(x)$ and $z' = \phi_{\omega}(x)$ and based on inequlity \ref{ineq: ftheta lipschitz} we get:

\begin{equation}\label{ineq: ftheta circ lipschitz }
\lVert f_{\theta}\circ \phi(x) - f_{\theta}\circ \phi_{\omega}(x) \lVert_2^2 \leq L^2 \lVert \phi(x) - \phi_{\omega}(x) \lVert_{2}^2
\end{equation}


Integrating both sides of inequality \ref{ineq: ftheta circ lipschitz } over $\mathcal{X}$ w.r.t $\mu$ we get: (note that integration is permitted since we have already established that all functions are p-integrable w.r.t. $\mu$)

\begin{equation}\label{ineq: fcircphi integral}
    \int_{x \in \mathcal{X}} \lVert f_{\theta}\circ \phi_{\omega}(x) - f_{\theta}\circ \phi(x) \lVert^2_2 \,d\mu(x) \leq L^2 \int_{x \in \mathcal{X}} \lVert \phi(x) - \phi_{\omega}(x) \lVert^2_2 \,d\mu(x) 
\end{equation}

which can be written as:

\begin{equation}\label{ineq: fcircphi integral norm}
    \lVert f_{\theta}\circ \phi_{\omega} - f_{\theta}\circ \phi \lVert_{L^2_{\mu}} \leq L^2 \lVert \phi_{\omega} - \phi \lVert_{L^2_{\mu}} 
\end{equation}

Note that the above inequality, as well as inequality \ref{ineq T-fcircphiomega}, holds for any neural network $\phi_{\omega}: \mathcal{X} \longrightarrow \mathcal{Z}$ with the above-mentioned activation functions.
In the Proof of Theorem 1 in \citet{korotin2023neural_OT}, 
it is shown that neural networks of the aforementioned form 
will be dense in $L_{\mu}^2(\mathcal{X},\mathcal{Z})$. As such, for convenience, we can choose a bound $\epsilon/{2L^2}$ s.t for any $\epsilon$, there exists a neural network $\phi_{\omega}$ s.t:

\begin{equation}\label{ineq: phi-omega - phi}
    \lVert \phi_{\omega} - \phi \lVert_{L^2_{\mu}} \leq \epsilon/{2L^2}
\end{equation}

Combining inequalities \ref{ineq: fcircphi integral norm}, \ref{ineq: phi-omega - phi} and \ref{ineq T-fcircphiomega} we get:

\begin{equation}\label{ineq: phi-omega - phi e/2}
     \lVert T - f_{\theta}\circ \phi_{\omega}\lVert \leq \epsilon
\end{equation}

which holds for any $\epsilon >0$, thus concluding the proof.


\section{Experimental details}\label{app: experiments}

\textbf{Neural networks.} We use vanilla MLPs with hidden sizes [128, 64, 64] and a ReLU activation function for all perametrizations, i.e for all $\phi_\omega,\widetilde{T}_\theta, {T}_\theta'$.
Specifically for the isomorphism network $\phi_\omega$ we add a residual connection from its first to its last layer. Inspired by the initialization scheme in \citet{gap_monge}, this approach encourages the network to learn an affine transformation between the source and reference samples. We initialize all networks using the orthogonal initialization, i.e which uses uniformly distributed orthogonal matrices. We train $\phi_\theta$ with a learning rate of $\eta = 10^{-3}$ and all other networks with $\eta = 10^{-4}$. In all cases, we train our neural networks using the ADAM optimizer \citep{kingma2017adam}, with a batch size of $n=1024$.

\textbf{Loss functions.} For the training of $\phi_\omega$, both during its pre-training phase and during the training of the composition map, we use the Sinkhorn divergence with the squared Euclidean as its fitting loss, i.e $\mathcal{S}_{l_2^2} = \Delta(\phi_\omega \sharp \hat{\mu}_n, \hat{\rho}_n)$, with an etropic regularization parameter of $\varepsilon = 0.01$. When training networks $\widetilde{T}_\theta$ and ${T}_\theta'$, we use the entropic Wasserstein distance \citep{cuturi2013sinkhorn} with the Euclidean distance as the cross-domain cost. For all fitting losses, we scale the cross-domain cost matrix using mean scaling. For all regularizers $\mathcal{GM}^{2}(\hat{\mu}_{n},\phi_{\omega} \sharp \hat{\mu}_{n})$, $\mathcal{GM}^2(\hat{\rho}_n,\widetilde{T}_{\theta} \sharp \hat{\rho}_n)$, $\mathcal{GM}^2(\hat{\mu}_n,{T}_{\theta}' \sharp \hat{\mu}_n)$ we use the quadratic entropic Gromov-Wasserstein distance \citep{peyre2016}. We use the Euclidean distance for all inter-domain costs $c_\mathcal{X}=c_\mathcal{Y}=c_\mathcal{Z} := \lVert \cdot \lVert_2$ and scale them using max scaling. 
We use a regularization strength of $\lambda_{GM} = 1$ across all losses.
For all fitting losses and regularizers, we use an entropic regularization parameter of $\varepsilon = 0.001$.


\end{document}